\documentclass[review,5p,twocolumn]{elsarticle}
\usepackage{amsmath}
\usepackage{bm}
\usepackage{graphicx}
\usepackage{colortbl}  
\usepackage{setspace} 
\usepackage{nicefrac}       
\usepackage{amsfonts} 
\usepackage{algorithm}
\usepackage{algorithmicx} 
\usepackage{algpseudocode}
\usepackage{tabularx}
\usepackage{wrapfig}
\usepackage{subcaption}
\usepackage{caption}
\usepackage{booktabs}
\usepackage[table,xcdraw]{xcolor}
\usepackage{hyperref}
\usepackage{multirow}
\usepackage{stfloats}

\journal{Journal of \LaTeX\ Templates}

\usepackage{placeins}

\begin{document}
\begin{frontmatter}

\title{SeaMo: A Season-Aware Multimodal Foundation Model for Remote Sensing}


\author[mymainaddress,mysecondaryaddress]{Xuyang Li}

\author[thirdaddress]{Chenyu Li}

\author[fourthaddress]{Gemine Vivone}

\author[mymainaddress,mysecondaryaddress,fifthaddress]{Danfeng Hong
}
\ead{hongdf@aircas.ac.cn}

\address[mymainaddress]{Aerospace Information Research Institute, Chinese Academy of Sciences, Beijing, 100094, China}
\address[mysecondaryaddress]{School of Electronic, Electrical and Communication Engineering, University of Chinese Academy of Sciences, Beijing, 100049, China}
\address[thirdaddress]{School of Mathematics and Statistics, Southeast University, Nanjing, 211189, China}
\address[fourthaddress]{Institute of Methodologies for Environmental Analysis, National Research Council, CNR-IMAA, Tito, 85050, Italy}
\address[fifthaddress]{ Graduate School of Frontier Sciences, the University of Tokyo, Chiba 277-8561, Japan}

\begin{abstract}
Remote Sensing (RS) data encapsulates rich multi-dimensional information essential for Earth observation. Its vast volume, diverse sources, and temporal continuity make it particularly well-suited for developing large Visual Foundation Models (VFMs). These models serve as powerful feature extractors, leveraging extensive RS data for pretraining and subsequent fine-tuning in various geoscientific applications. However, existing VFMs in the RS domain often concentrate on specific image characteristics, neglecting the full season-aware potential of RS data. To bridge this gap, we introduce SeaMo, a novel VFM that effectively integrates multimodal and multi-seasonal RS information. SeaMo leverages a masked image modeling framework to fully exploit the spatial, spectral, and seasonal dimensions of RS data. Specifically, we employ unaligned spatial region selection to capture spatial heterogeneity, incorporate multi-source inputs for enhanced multimodal integration, and introduce temporal-multimodal fusion blocks to assimilate seasonal variations effectively. By explicitly modeling the complex, season-dependent attributes of RS data, SeaMo enhances generalization, robustness, and adaptability across geoscientific tasks. Extensive experiments and ablation studies demonstrate its superior performance, underscoring its potential as a foundational model for Earth observation.
\end{abstract}


\end{frontmatter}


\section{Introduction}
\label{Introduction}
Remote sensing is a technology that enables the measurement of soil or crop characteristics using platforms such as Unmanned Aerial Vehicles (UAVs), airplanes, or satellites. Data acquired through remote sensing technology are utilized to study social phenomena, land use dynamics, and changes in landscapes for scientific analysis and modeling purposes~\cite{vivone2024deep}. Currently, hundreds of remote sensing satellites continuously monitor the surface of the Earth, generating large-scale time series datasets. These datasets, characterized by their heterogeneity and incorporation of multiple sources, including spectral data, radar data, and meteorological data, provide a rich repository of physical geographic information, embodying the typical traits of big data~\cite{hong2024spectralgpt}. The research on image data in remote sensing currently faces several challenges and focal points:

\textbf{(1) Integration of data from diverse sources for varied geoscientific applications}: Different methods of capturing how electromagnetic radiation interacts with Earth's surface materials provide unique data sets~\cite{li2024casformer}. Optical data, by measuring radiation across many wavelengths, offers detailed spectral information that characterizes the material composition of objects. Synthetic Aperture Radar (SAR) data, by transmitting and receiving longer wavelength electromagnetic pulses, can assess the geometry, roughness, and electrical properties of objects. Given the significant heterogeneity between these data sources, simple transformations and combinations of data dimensions are insufficient to fully exploit this information~\cite{hong2024multimodal}. Thus, a key area of focus is how to extract valuable geoscientific information from these diverse datasets that can meet human needs.

\textbf{(2) Effective modeling and utilization of spatio-temporal data with seasonal awareness:}
Satellite imagery inherently possesses a spatio-temporal structure as it consists of multiple observations of the same location over time. This rich temporal information enables the analysis of changes in physical geography, land use, and even offers insights into geological structures or socio-economic conditions~\cite{xu2023ai}. However, the low information density resulting from low spatial resolution and long intervals between observations poses challenges for capturing temporal dynamics with simple techniques. Moreover, many existing methods overlook the critical impact of seasonal variations. Recognizing that seasonal changes play a fundamental role in shaping remote sensing observations, it becomes essential to develop models that are explicitly season-aware.

VFMs refer to large-scale neural networks that are pretrained on vast datasets, typically using self-supervised learning methods. VFMs act as robust and powerful feature extractors and can be fine-tuned with minimal effort for various downstream tasks~\cite{bommasani2021opportunities}. Given the inherent big data characteristics and the complex information contained within remote sensing data, the vast data landscape of remote sensing offers significant opportunities for the implementation of VFMs in Earth observation. Currently, several VFMs have been introduced in the field of remote sensing, demonstrating performance that surpasses traditional deep learning models on a broad range of geoscientific tasks. This advancement has led to significant breakthroughs in the mining and extraction of remote sensing information. However, these models still face several challenges and deficiencies:

\textbf{(1) Lack of geoscientific attributes:} Some remote sensing VFMs are derived from computer vision technologies and often involve simplistic transformations of remote sensing data to adapt it to algorithmic requirements. This approach generally leads to insufficient exploration of the geoscientific properties inherent in remote sensing data, resulting in a gap in feature extraction capabilities. For instance, DINO-MC~\cite{wanyan2024extending} employs basic data augmentation and uses Contrastive Learning (CL) for pretraining. Similarly, RSVA~\cite{wang2022advancing} straightforwardly adapts the Masked AutoEncoders (MAE) from the computer vision domain for pretraining in remote sensing.

\textbf{(2) Single-dimension data modeling:} Although most remote sensing VFMs acknowledge the unique characteristics of remote sensing data, they typically focus on a single dimension, such as spatial or temporal attributes~\cite{li2024s2mae}. For instance, SeCo~\cite{manas2021seasonal} and CaCo~\cite{mall2023change} exploit the temporal dynamics of RS images, employing CL to capture time-invariant features at specific geo-locations. RingMo~\cite{sun2022ringmo} introduces Masked Image Modeling (MIM) to the RS domain, targeting small objects with a sparse masking strategy. SatMAE~\cite{cong2022satmae} employs a group combination masking strategy within the MAE framework~\cite{he2022masked}, enriching its handling of multi-spectral or temporal data. Similarly, SpectralGPT~\cite{hong2024spectralgpt} innovatively addresses the spectral sequence attributes of images, employing a 3D mask strategy focused on the reconstruction of spectral sequences. ScaleMAE~\cite{reed2023scale} and SatMAE++~\cite{noman2024rethinking} adjust to the varying sizes and resolutions of RS images, designing multiple image-size reconstruction targets. CROMA~\cite{fuller2024croma} pretrains aligned radar and optical images using both CL and MAE methods.

\textbf{(3) Non-explicit modeling multi-dimensional data attributes:} Some remote sensing VFMs consider multi-dimensional attributes, but their approach to modeling and pretraining is more akin to stacking and assembling rather than explicitly modeling from a remote sensing and geoscientific perspective~\cite{guo2023skysense,bastani2023satlaspretrain}. This often leads to a superficial integration of multidimensional data, which does not fully exploit the synergistic potential of combining spatial, temporal, and spectral characteristics inherent in remote sensing data. For instance, Skysense~\cite{guo2023skysense} integrates information from various RS data modalities, temporal information, and geographic location within the framework of CL. It employs CL for multimodal spatial feature extraction and utilizes positional encodings to embed temporal and geographic information, surpassing previous models. However, the integration of its multi-dimensional attributes is not fully realized. Specifically, the model optimizes the data learning from multiple separate aspects rather than in a holistic manner. The embedding of positional encodings serves merely as an insertion of prior knowledge rather than facilitating active knowledge learning and discovery.

Motivated by these challenges, we argue that it is essential to develop a VFM that not only leverages the extensive, multi-source satellite data but also explicitly models both spatial and temporal properties under a unified training objective. To this end, we introduce \textbf{SeaMo}, a season-aware multimodal foundation model for remote sensing that effectively integrates spatial, multimodal, and multi-seasonal information. First, to enhance the spatial information density, we employ unaligned spatial region selection techniques. Importantly, these techniques are not merely used as data augmentation tools; they play a critical role in promoting the discovery of spatially diverse features and in challenging the model to learn more robust representations. Second, to harness the complementary strengths of multimodal satellite data during pretraining, we utilize a unified encoder that jointly accepts multisource data, with integration facilitated via self-attention mechanisms. Third, to improve time-invariant representations, we design a temporal fusion block that leverages cross-attention mechanisms. This block is more than a standard cross-attention module; it is purpose-built to fuse multiple spatial-temporal data streams, thereby capturing intricate relationships across both time and modality. Finally, the model is optimized using a unified masked modeling approach. Our contributions are outlined as follows:
\begin{itemize}
\item We introduce SeaMo, a novel multimodal foundation model that leverages multi-season temporal and multimodal heterogeneous information embeddings. SeaMo integrates unaligned spatial region selection techniques to improve spatial diversity and model robustness effectively.
\item We design a temporal-multimodal fusion block, a cascade-style module that effectively integrates seasonal and multimodal information during pretraining. By employing a specialized form of cross-attention, this block goes beyond traditional mechanisms to capture intricate temporal-spatial interactions.
\item Our pretraining strategy adopts a multi-stage progression, evolving from unimodal learning to multimodal and ultimately seasonal-multimodal representation learning. This staged approach enables a more comprehensive understanding of the complex, multidimensional features characteristic of remote sensing data.
\item The proposed foundation model has been effectively transferred to a range of downstream geoscientific applications across diverse modalities, consistently demonstrating strong performance. Comprehensive ablation studies further validate its superior capabilities. By explicitly embedding seasonal awareness into multimodal feature extraction, SeaMo marks a substantial advancement in the development of foundation models for Earth observation.
\end{itemize}

\section{Related Work}
\label{Related_Work}
\subsection{Self-supervised visual representation learning}
Self-supervised learning is a technique for learning generalizable visual representations through various pretext tasks without the need for annotations during pretraining~\cite{bommasani2021opportunities}. Initially, this approach utilized straightforward tasks such as image colorization, image rotation prediction, or solving jigsaw puzzles~\cite{srivastava2015unsupervised, vondrick2016anticipating}. In recent years, CL and MIM have become predominant self-supervised techniques for pretraining large foundation models. CL is designed to capture feature invariance in a batch of images by differentiating between identical and modified images using data augmentation techniques, with implementations like SimCLR~\cite{chen2020simple} and MoCo~\cite{he2020momentum}. In contrast, MIM focuses on revealing spatial correlations by reconstructing masked patches within a single image, as seen in methods such as SimMIM~\cite{xie2022simmim} and MAE~\cite{he2022masked}. These techniques are complementary. Methods such as SiameseIM \cite{tao2023siamese} successfully integrate CL and MIM to capture data features, leading to superior performance on both linear-probing and fine-tuning tasks.
\begin{figure*}[t]
      \centering	   
      \includegraphics[width=1\textwidth]{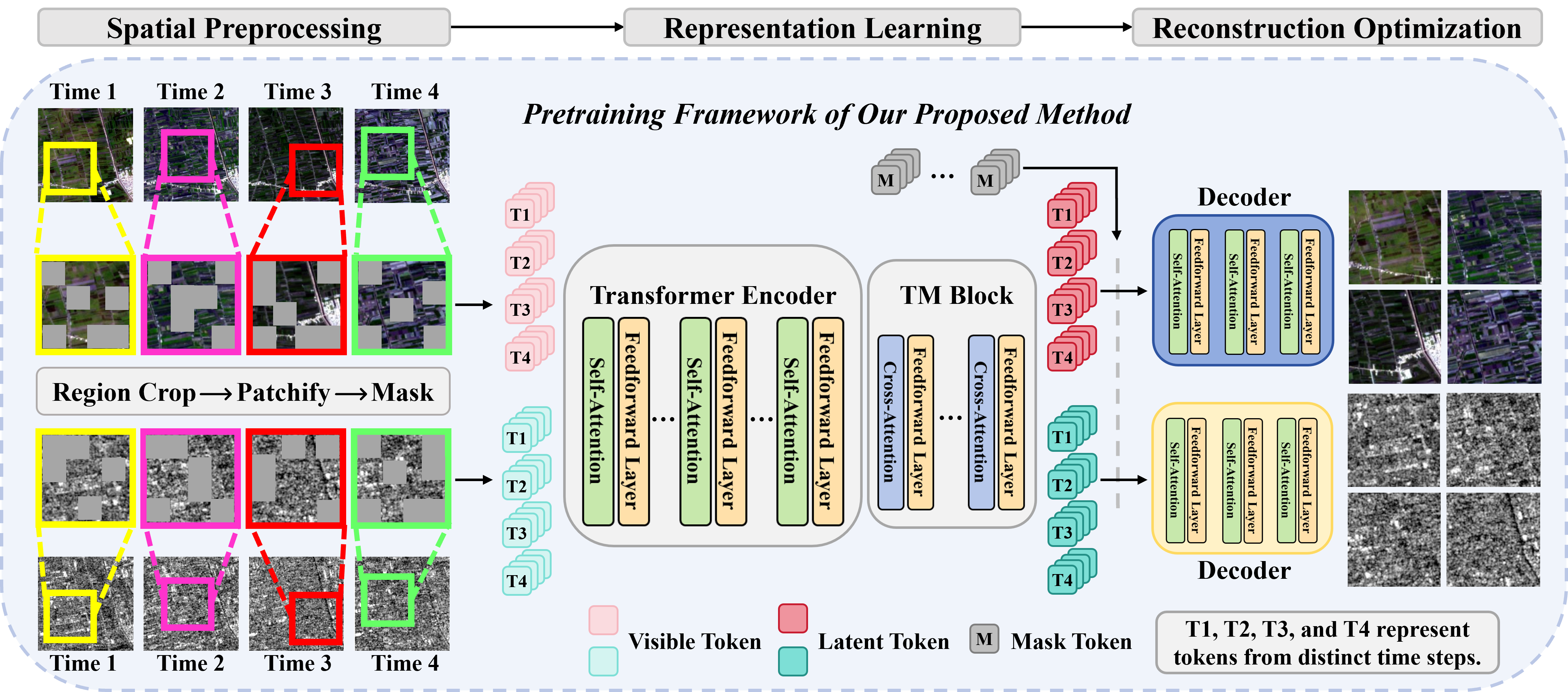}
      \caption{\textbf{Pretraining workflow of the SeaMo foundation model.} The SeaMo architecture integrates three primary components: encoders, Temporal-Multimodal fusion blocks (TM blocks), and decoders. Our approach incorporates a partially overlapping spatial selecting strategy, ensuring that images from the same temporal instance are selected identically across various modalities, while images from different instances exhibit partial overlaps. These processed images then serve as inputs to the network. Following the masked autoencoder paradigm, only visible tokens are processed by the encoder. The TM block effectively merges features from multiple seasons and modalities, culminating in modality-specific decoders that reconstruct the initially masked regions of the images.}
      
\label{fig:workflow}
\end{figure*}
\subsection{Masked autoencoders}
MAEs~\cite{he2022masked} are a variant of denoising autoencoders that learn representations by reconstructing the original image from its masked inputs. It introduces random masks into the image inputs, requiring the network to reconstruct the original image from a partially masked version as a means of pretraining on image data. Typically, the model employs an encoder-decoder structure, and after pretraining, only the encoder is used as the backbone for downstream tasks. In MIM, the encoder-decoder structure is usually asymmetric: the encoder is designed to learn high-level representations, while the decoder focuses on learning how to reconstruct the target. MAEs have demonstrated their effectiveness on a wide range of vision benchmarks, prompting their adaptation in numerous studies spanning various data modalities. These include multimodal images~\cite{bachmann2022multimae}, video data~\cite{feichtenhofer2022masked}, medical data~\cite{zhou2023self}, remote sensing data~\cite{li2024s2mae} and meteorological data~\cite{nguyen2023climax}.
\begin{figure}[t]
      \centering	   
      \includegraphics[width=0.48\textwidth]{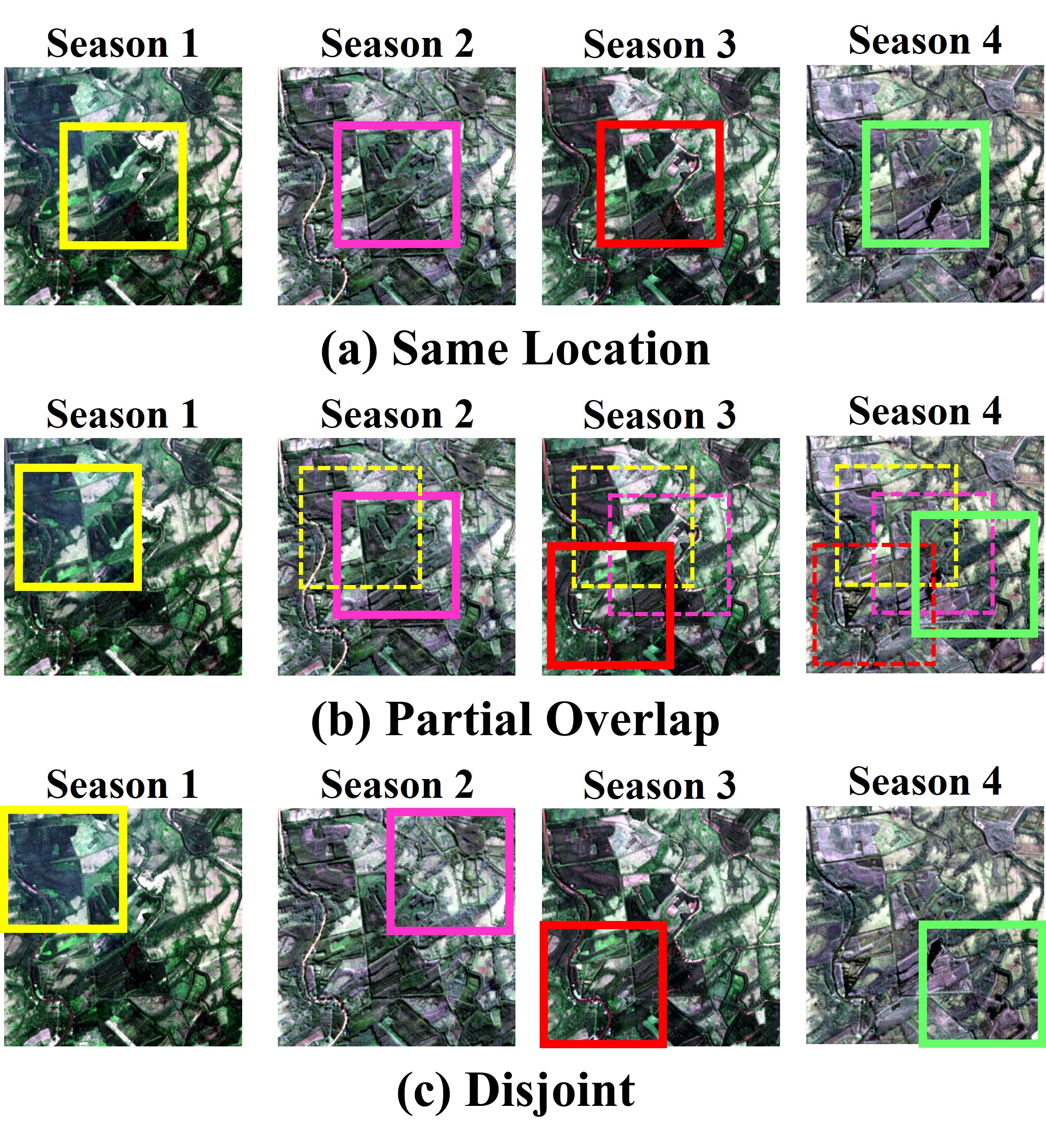}
      \caption{\textbf{Different region selection strategies for temporal data.} The solid boxes indicate the image regions that are selected and fed into the network. (a) Images from different seasons are selected from the same section. (b) Images from different seasons are selected based on a specific proportion of the full image, ensuring partial overlap. (c) Images from different seasons are selected with no overlap.}
\label{fig:crop}
\end{figure}

\subsection{Remote sensing foundation models}
As foundational models in remote sensing continue to flourish, new directions are emerging in remote sensing image processing. In this field, the training data for foundational models has expanded from single-sensor data to multi-sensor and even multi-modal data. For example, GFM~\cite{mendieta2023towards} and ScaleMAE~\cite{reed2023scale} pretrain on high-resolution RGB images, while Cross-Scale MAE~\cite{tang2023cross} combines CL with MAE for training on high-resolution RGB images. In contrast, Models like SpectralGPT~\cite{hong2024spectralgpt} develop self-supervised strategies tailored to the spectral characteristics of multispectral imagery. Additionally, some models enhance their training data sources by embedding extensive geographical and temporal information~\cite{ayush2021geography,10726860}. For instance, Skysense~\cite{guo2023skysense} integrates location and temporal cues into a multimodal CL framework, whereas CaCo~\cite{mall2023change} performs inter-temporal CL to significantly raise the performance ceiling. 
In the development of Vision-Language Models (VLMs), the primary focus has been on aligning text with remote sensing images to enhance image understanding and facilitate interaction between remote sensing images and Large Language Models (LLMs). Models such as RemoteCLIP~\cite{10504785} and GeoCLIP~\cite{10.5555/3666122.3666501} achieve few-shot capabilities, while some VLMs fine-tune both VLMs and LLMs on remote sensing-specific instruction datasets, enabling these models to interpret remote sensing images and engage in multi-turn visual dialogue tasks~\cite{li2024vision}. The model proposed in this work specifically targets the explicit fusion of data from multiple sensor sources while embedding temporal information.

\section{Proposed Method}
\label{headings}
We propose \textbf{SeaMo}, a novel season-aware multimodal foundation model for remote sensing that learns robust spatial and temporal representations from multi-seasonal, multisource data. SeaMo comprises both a deep network architecture and an integrated training strategy. The network architecture includes a \emph{Multimodal Patch Embedding Module} that transforms raw optical and SAR images into token embeddings, a \emph{Unified Multimodal Feature Encoder} that jointly accepts multi-source data and integrates it via self-attention mechanisms, a \emph{Temporal Fusion Module for Multimodal Integration} that leverages cross-attention to fuse information across seasons and modalities, and \emph{Modality-Specific Reconstruction Modules} that recover the original images through masked image modeling. Complementing this architecture, our training strategy incorporates \emph{Geospatial Region Selection Strategies} to enhance spatial representation learning via diverse cropping methods, as well as a \emph{Progressive Pretraining Strategy} that transitions the model from single-time point multimodal learning to a comprehensive multi-time flow framework. Figure~\ref{fig:workflow} illustrates the overall SeaMo framework. Our approach is pretrained on the SSL4EO-S12 dataset~\cite{wang2023ssl4eo}, which includes 12-channel multispectral optical imagery from Sentinel-2 and 2-channel SAR backscatter data from Sentinel-1.

\subsection{SeaMo overview}
\label{Method_overview_of_SeaMo}

\subsubsection{Multimodal patch embedding module}
We denote the optical inputs as 
\(
\bm{I}_O \in \mathbb{R}^{T \times C_O \times H \times W},
\)
and the SAR inputs as 
\(
\bm{I}_R \in \mathbb{R}^{T \times C_R \times H \times W}.
\)
Here, \(T\) represents the number of temporal instances (or seasons), \(C_O\) and \(C_R\) denote the number of channels for optical and SAR data respectively, \(H\) and \(W\) are the height and width of each image. For each season \(t = 1, 2, \ldots, T\), the corresponding images are given by
\(
\bm{I}_O^t \in \mathbb{R}^{C_O \times H \times W} \quad \text{and} \quad \bm{I}_R^t \in \mathbb{R}^{C_R \times H \times W}.
\)
Each modality-specific image is processed by its dedicated embedding module, which converts the image into a set of \(L\) patch tokens. Specifically, for each season \(t\), the embedding modules yield token encodings
\(
\bm{E}_O^t, \ \bm{E}_R^t \in \mathbb{R}^{L \times D},
\)
where \(L\) is the number of tokens and \(D\) is the embedding dimension. Fixed sinusoidal positional embeddings~\cite{vaswani2017attention} are added to these token encodings to preserve spatial order.

\subsubsection{Selective masking of patch tokens}
A selective masking operation is applied to the token embeddings to differentiate between visible and masked patches. This operation is defined as
\(
[\bm{O}^t_{\text{visible}}, \bm{O}^t_{\text{masked}}] = \mathbb{M} \odot \bm{E}_O^t \quad \text{and} \quad [\bm{R}^t_{\text{visible}}, \bm{R}^t_{\text{masked}}] = \mathbb{M} \odot \bm{E}_R^t,
\)
where \(\odot\) denotes element-wise multiplication and \(\mathbb{M} \in \{0,1\}^{L \times D}\) is a binary mask specifying which tokens are to be masked. Notably, a consistent masking strategy is applied for images from different sensors captured simultaneously, while a random masking strategy is adopted for images acquired at different times. This dual strategy challenges the model to learn more robust and discriminative representations.

\subsubsection{Unified multimodal feature encoding}
We concatenate the visible tokens from both modalities to form a unified token set for each season \(t\). Specifically, given the optical tokens 
\(
\bm{O}^t_{\text{visible}} \in \mathbb{R}^{L \times D}
\)
and the SAR tokens 
\(
\bm{R}^t_{\text{visible}} \in \mathbb{R}^{L \times D},
\)
we create the combined input:
\(
X^t = \text{concat}\left[\bm{O}^t_{\text{visible}}, \bm{R}^t_{\text{visible}}\right] \in \mathbb{R}^{2L \times D},
\)
where \(\text{concat}[\cdot,\cdot]\) denotes concatenation along the token dimension. This combined token set \(X^t\) is then fed into the encoder \(\Phi\), which is composed of standard vision transformer blocks~\cite{dosovitskiy2020vit}. Within each transformer block, self-attention is computed as follows:
\[
\text{Attention}(Q, K, V) = \text{softmax}\left(\frac{QK^\top}{\sqrt{D}}\right)V,
\]
where the query, key, and value matrices are obtained by applying learned linear projections on \(X^t\):
\(
Q = X^tW_Q, K = X^tW_K, V = X^tW_V,
\)
with \(W_Q\), \(W_K\), and \(W_V\) being the projection matrices.

Through the self-attention mechanism, the encoder effectively captures the inter-token relationships across both modalities and spatial dimensions. As a result, the encoder outputs multimodal features denoted as
\(
\{\bm{F}^t_{O}, \bm{F}^t_{R}\} = \Phi(X^t),
\)
which encapsulate the rich spatial and modality-specific characteristics present in the input data.

\begin{figure*}[ht]
      \centering	   
      \includegraphics[width=0.9\textwidth]{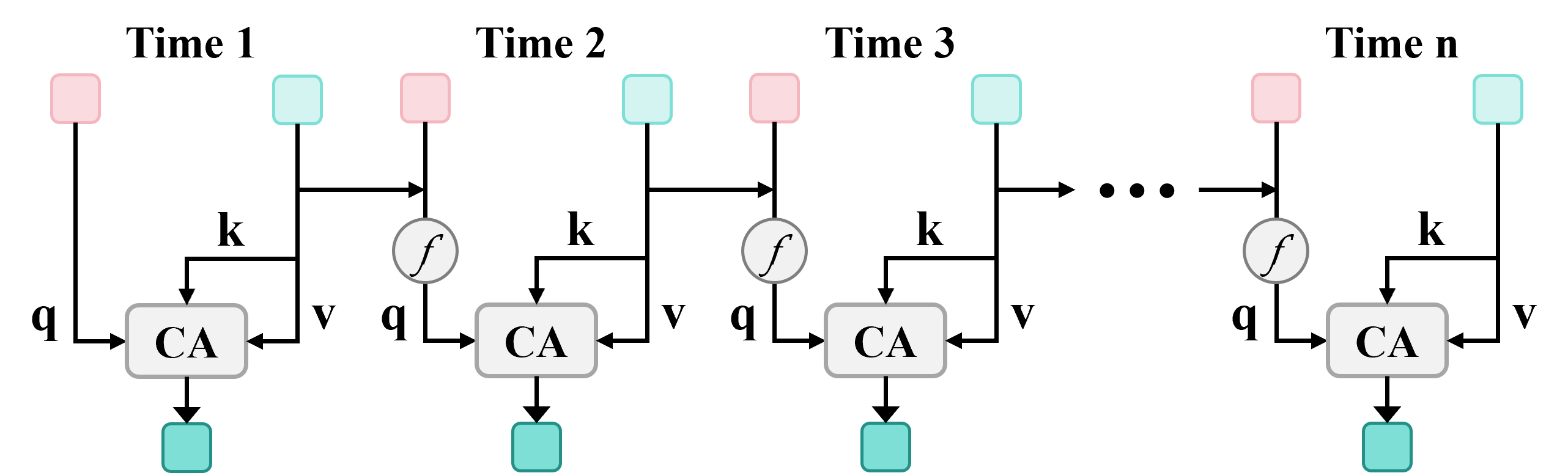}
      \caption{\textbf{An illustration of the Temporal-Multimodal (TM) block.} In this block, data from each modality not only participate in fusion interactions during the current season but also influence the fusion process in subsequent seasons. For clarity, the symbols in the figure are defined as follows: $k$ denotes the key vector, $q$ denotes the query vector, and $v$ denotes the value vector; $CA$ represents cross-attention; and $f$ indicates the fully connected layer.}
\label{fig:tm_block}
\end{figure*}

\subsubsection{Temporal fusion module for multimodal integration}
Although the encoder processes seasonal images to yield multimodal features \(\{\bm{F}^t_{O}, \bm{F}^t_{R}\}\) for \(t=1,2,\ldots,T\), it does not explicitly capture temporal dependencies. To remedy this, we propose the Temporal-Multimodal (TM) fusion block (see Figure \ref{fig:tm_block}), inspired by SiamMAE~\cite{gupta2023siamese}. Unlike traditional cross-attention mechanisms that typically focus on interactions along a single dimension (e.g., between queries and keys from different modalities), the TM block is specifically designed to fuse both temporal and modal information concurrently. In this block, data from each modality not only participate in the fusion process at the current time point but also influence the fusion at subsequent time points. 

In practice, the visible tokens of each image serve as queries. Tokens from the other modality of the same season, together with tokens from the previous season, are utilized as keys and values. This joint use of tokens from different modalities and temporal instances enables the TM block to establish cross-attention relationships that integrate both temporal evolution and multimodal interactions, thereby leading to richer and more season-aware representations than conventional cross-attention alone. Algorithm~\ref{algorithm1} provides the pseudocode for the TM fusion block.

\subsubsection{Modality-specific reconstruction modules}
The output of the TM fusion block consists of refined features for each modality at every season, denoted as \(\{\bm{H}^t_{O}, \bm{H}^t_{R}\}\) for \(t=1,2,\ldots,T\). For each modality, we design dedicated reconstruction modules that share weights across seasons. Specifically, the optical features \(\bm{H}^t_{O}\), augmented with mask tokens and positional embeddings, are input to the optical reconstruction module to regenerate the original images \(\bm{I}_O^t\). The SAR modality follows a similar reconstruction process. The reconstruction loss is computed as the Mean Squared Error (MSE) between the reconstructed and original images, calculated only for the masked patches. The overall loss is obtained by summing the reconstruction losses across all seasons and modalities.
\begin{algorithm*} 
\setstretch{1.2} 
\caption{Temporal-Multimodal fusion block}
\begin{algorithmic}[1] 
\State \textbf{Input:} Encoded features \(\mathbf{F}^t_{O}, \mathbf{F}^t_{R}\) for \(t = 1, 2, \ldots, T\)
\State \textbf{Output:} Temporally and modally fused features \({\mathbf{H}}^t_{O}, {\mathbf{H}}^t_{R}\) for \(t = 1, 2, \ldots, T\)

\For{\(t = 1\) to \(T\)}
    \State \textbf{Multimodal fusion at time \(t\)}
    \State \({\mathbf{H}}^t_{O} \gets CA(\text{query} = \mathbf{F}^t_{R}, [\text{key, value}] = \mathbf{F}^t_{O})\)
    \State \({\mathbf{H}}^t_{R} \gets CA(\text{query} = \mathbf{F}^t_{O}, [\text{key, value}] = \mathbf{F}^t_{R})\)
    \If{\(t < T\)}
        \State \textbf{Temporal fusion for next time point}
        \State \({\mathbf{H}}^{t+1}_{O} \gets CA(\text{query} = f[\mathbf{F}^{t+1}_{R}, \mathbf{F}^t_{O}], [\text{key, value}] = \mathbf{F}^{t+1}_{O})\)
        \State \({\mathbf{H}}^{t+1}_{R} \gets CA(\text{query} = f[\mathbf{F}^{t+1}_{O}, \mathbf{F}^t_{R}], [\text{key, value}] = \mathbf{F}^{t+1}_{R})\)
    \EndIf
\EndFor
\State \textbf{Note:} \(CA\) denotes the cross-attention layer and \(f\) represents the fully connected layers used to fuse features.
\end{algorithmic}
\label{algorithm1}
\end{algorithm*}
\subsection{Geospatial region selection strategies}
\label{crop_strategies}
Despite the rich geospatial information contained in remote sensing images, their inherent low spatial resolution and expansive coverage yield a low information density at the image level. As a result, relying solely on image-level self-supervised learning can impede the capture of fine-grained spatial details. In contrast, temporal data provide repeated observations of the same geographical area over different periods, thereby offering an opportunity to enhance spatial representation learning via diverse cropping strategies. For instance, one strategy employs an identical crop region for all temporal images, as illustrated in Figure \ref{fig:crop}(a). Alternatively, crops may be selected to ensure partial spatial overlap among temporal images, as demonstrated in Figure \ref{fig:crop}(b). In the most challenging scenario, crops from different temporal images are entirely non-overlapping, as depicted in Figure \ref{fig:crop}(c).

From a remote sensing perspective, we argue that, aside from changes induced by human activities, the geological attributes of a given area tend to remain consistent over time. By enforcing partially overlapping cropping on same-modality images across different time points, the network is encouraged to learn correlations even in non-overlapping regions. This strategy is conceptually similar to the approach adopted by CropMAE~\cite{eymael2024efficient}, which demonstrated that employing locally correlated cropping in MAE tasks can substantially enhance spatial representation learning. However, while CropMAE primarily focuses on single-image spatial consistency, our method leverages the temporal dimension to further constrain the network and uncover latent cross-temporal correlations.
In Section \ref{Ablation_studies_of_Crop_strategies}, we evaluate the three cropping methods illustrated in Figure \ref{fig:crop}. Our experiments indicate that partially overlapping data inputs yield the best performance. We believe that using the identical region for all temporal images oversimplifies the reconstruction task, whereas using entirely non-overlapping crops makes it overly challenging for the network to discover inter-region correlations.

\subsection{Progressive pretraining strategy}
\label{Progressive_pretraining_strategy}
As depicted in Figure \ref{fig:con_pretraining} (left), solely using a single MIM model in a unimodal pretraining framework restricts cross-modal interactions, resulting in implicit and often suboptimal multimodal representations. In contrast, we employ the progressive pretraining strategy, which not only leverages the inherent strengths of spatial and temporal data but also introduces a novel two-phase training scheme that distinctly sets our work apart from previous approaches such as GFM~\cite{mendieta2023towards}, SpectralGPT~\cite{hong2024spectralgpt}, and other MAE-based methods~\cite{xiong2024neural}. By sequentially focusing on single-time point features and then on multi-time flow multimodal learning, our strategy delivers several key advantages:

\begin{itemize}
    \item \textbf{Enhanced spatial detail capture:} The first phase concentrates on learning fine-grained spatial and multimodal features from a single time point, laying a robust foundation that directly addresses the challenges posed by low spatial resolution and sparse information density in remote sensing images.
    \item \textbf{Effective temporal fusion:} The second phase explicitly integrates temporal information through TM blocks, enabling the model to fuse cross-temporal data via cross-attention mechanisms. This design is not only novel but also crucial for capturing time-invariant representations, which are often overlooked in conventional MIM frameworks.
    \item \textbf{Improved representation quality:} By progressively refining the model's capabilities, from basic spatial understanding to complex temporal integration, our approach achieves a more robust and comprehensive feature representation than methods relying on a single-stage pretraining process.
\end{itemize}
\subsubsection{Single-time point multimodal learning}
\begin{figure*}[t]
      \centering	   
      \includegraphics[width=0.9\textwidth]{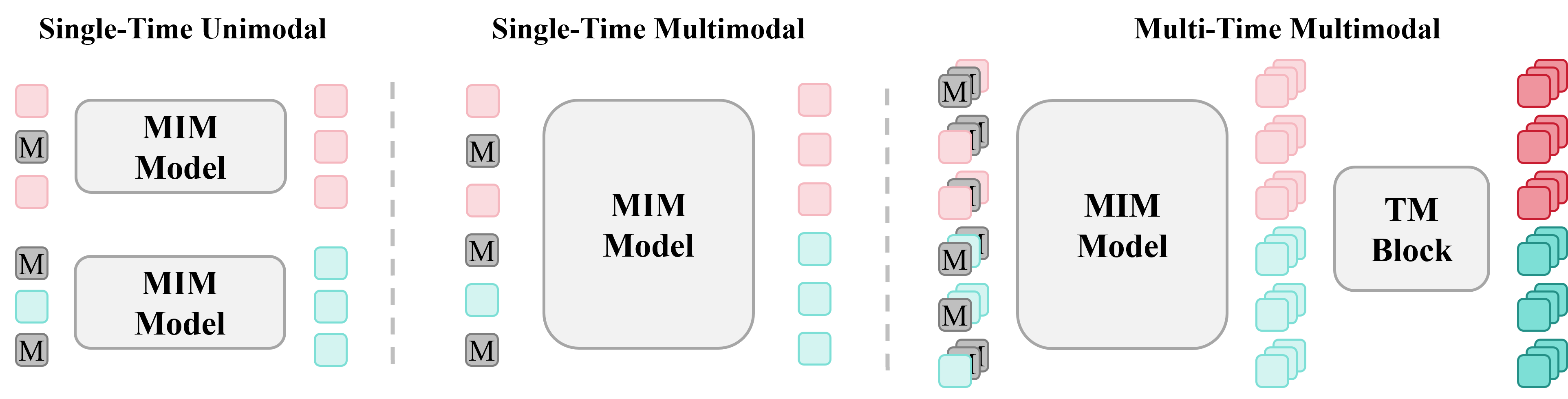}
      \caption{\textbf{Three distinct multimodal pretraining strategies.}
\textbf{Left}: The MIM model is weight-sharing, however, lacks interaction across different modalities.
\textbf{Middle}: Data from two modalities are concatenated and then fed into the MIM model.
\textbf{Right}: A time series of multimodal images are concatenated (modal-concat) and then processed by the MIM model, followed by a TM block to strengthen representation learning.} 
\label{fig:con_pretraining}
\end{figure*}

As illustrated in Figure \ref{fig:con_pretraining} (middle), in the initial phase, we focus on learning high-quality spatial and multimodal features from a single temporal snapshot. Specifically, we pretrain the model using only the first seasonal dataset, which accounts for 25\% of the total data and intentionally excludes temporal information. During this stage, visible tokens from the two modalities are concatenated and fed directly into the encoder. This approach emphasizes the extraction of fine-grained spatial details and cross-modal correlations, which are critical for remote sensing applications due to the low information density of individual images. By concentrating on a single time point, the model benefits from a simplified training scenario that encourages the learning of robust spatial representations. This foundational stage not only stabilizes the initial learning process but also lays the groundwork for integrating temporal dynamics in the subsequent phase.
\subsubsection{Multi-time flow multimodal learning}
In this stage, we initialize the model using pretrained weights from the single-time learning stage. The inputs for this phase include multi-seasonal and multimodal images. Following initial processing by the encoder, the TM blocks are integrated to facilitate interactions among seasonal data, as depicted in Figure \ref{fig:con_pretraining} (right). Within these blocks, multi-seasonal and multimodal tokens are explicitly fused and interact through cross-attention layers in a time flow sequence. TM blocks are essential for enabling the model to acquire time-invariant multimodal representations, significantly enhancing model performance. This functionality has been validated through our ablation studies in Section \ref{ablation_studies}. After pretraining, the encoder serves as the backbone for downstream tasks, while the TM blocks and decoders are discarded.

\begin{table*}[ht]
\centering
\setlength{\tabcolsep}{2pt} 
\caption{\textbf{Fine-Tuning (FT) results on three optical classification benchmarks: EuroSAT,  fMoW-S2, and BigEarthNet.} On the fMoW-S2 and BEN datasets, we default to using only 10\% of the training data to fine-tune the models. * denotes the results obtained by fine-tuning with all training data, rather than only 10\% of the data. Best and second-best results are in red and blue, respectively.}
\label{opt_FT}
\begin{tabular}{c|c|ccccccccccc}
\toprule
 \multirow{3}{*}{Dataset} & \multirow{3}{*}{Metrics}& GASSL & SeCo & MAE & DINO & I-JEPA & SatMAE & SpectralGPT & Satlas & CROMA & DOFA & \multirow{2}{*}{Ours} \\ 
   & & ~\cite{ayush2021geography} & ~\cite{manas2021seasonal} & ~\cite{he2022masked} & ~\cite{caron2021emerging} & ~\cite{assran2023self} & ~\cite{cong2022satmae} & ~\cite{hong2024spectralgpt} & ~\cite{bastani2023satlaspretrain} & ~\cite{fuller2024croma} & ~\cite{xiong2024neural} &  \\ \cline{3-13}
 &  & ResNet50 & ResNet50 & ViT-S & ViT-S & ViT-B & ViT-B & ViT-B & Swin-B & ViT-B & ViT-B & ViT-B \\
\midrule
\midrule
EuroSAT&Acc.&96.96&97.23&98.78&99.01&99.20&99.20&99.21&98.97&99.22&\textbf{\textcolor{blue}{\text{99.30}}}&\textbf{\textcolor{red}{\text{99.37}}} \\ \hline
fMoW-S2&Acc.&50.69\textsuperscript{*}&51.65\textsuperscript{*}&51.79&52.79&53.54&57.20&55.28&57.95&54.47&\textbf{\textcolor{blue}{\text{58.10}}}&\textbf{\textcolor{red}{\text{58.25}}} \\ \hline
BEN&mAP&79.24&82.62&86.15&87.04&85.92&85.94&87.50&82.80&\textbf{\textcolor{blue}{\text{87.58}}}&\text{86.75}&\textbf{\textcolor{red}{\text{88.54}}} \\ \hline

\end{tabular}
\end{table*}

\begin{table*}[ht]
\centering
\setlength{\tabcolsep}{2pt} 
\caption{\textbf{Linear-Probing (LP) results on three optical classification benchmarks: EuroSAT,  fMoW-S2, and BigEarthNet (BEN).} On the fMoW-S2 and BEN datasets, we default to using only 10\% of the training data to fine-tune the models. Best and second-best results are in red and blue, respectively.}
\label{opt_LP}
\begin{tabular}{c|c|ccccccccccc}
\toprule
 \multirow{3}{*}{Dataset} & \multirow{3}{*}{Metrics}& GASSL & SeCo & MAE & DINO & I-JEPA & SatMAE & SpectralGPT & Satlas & CROMA & DOFA & \multirow{2}{*}{Ours} \\ 
  & & ~\cite{ayush2021geography} & ~\cite{manas2021seasonal} & ~\cite{he2022masked} & ~\cite{caron2021emerging} & ~\cite{assran2023self} & ~\cite{cong2022satmae} & ~\cite{hong2024spectralgpt} & ~\cite{bastani2023satlaspretrain} & ~\cite{fuller2024croma} & ~\cite{xiong2024neural} &  \\ \cline{3-13}
 &  & ResNet50 & ResNet50 & ViT-S & ViT-S & ViT-B & ViT-B & ViT-B & Swin-B & ViT-B & ViT-B & ViT-B \\
\midrule
\midrule
EuroSAT&Acc.&86.52&87.62&86.08&87.04&85.92&86.87&87.92&91.14&91.75&\textbf{\textcolor{blue}{\text{92.20}}}&\textbf{\textcolor{red}{\text{93.46}}}  \\ \hline
fMoW-S2&Acc.&33.82&34.41&27.69&32.64&32.35&35.17&35.80&37.81&\textbf{\textcolor{red}{\text{38.17}}}&37.51&\textbf{\textcolor{blue}{\text{37.95}}} \\ \hline
BEN&mAP& 76.41&77.97&75.94&81.58&80.80&79.36&81.05&82.11&\textbf{\textcolor{red}{\text{83.41}}}&82.45&\textbf{\textcolor{blue}{\text{82.73}}} \\ \hline
\end{tabular}
\end{table*}
\section{Experiments}
\label{Experiments}
In this section, we first describe the dataset utilized for pretraining and outline the pretraining setup for our model, SeaMo. Then we report and compare the performance of SeaMo and other models across a range of downstream tasks, including single/multi-label classification, semantic segmentation, and change detection involving various modalities. The results demonstrate SeaMo's exceptional performance and generalization capabilities. In addition, we present ablation studies that explore different model configurations, hyperparameter settings, and choices of pretraining strategies.
\subsection{Pretraining experimental setup}
\label{Pretraining_Experimental_Setup}
\subsubsection{Pretraining data}
We pretrained our model SeaMo on the SSL4EO-S12~\cite{wang2023ssl4eo} dataset, a comprehensive multimodal, multi-temporal dataset designed for self-supervised learning in Earth observation. This dataset targets diverse urban and suburban landscapes and excludes images with high cloud cover. It includes 3 million images from Sentinel-2 (multi-spectral, levels 1C and 2A) and Sentinel-1 (SAR), covering 250,000 locations worldwide. Each location is documented with four seasonal snapshots, capturing the variability within a calendar year. The images are uniformly sized at $264\times264$ pixels. For pretraining, we utilize products from Sentinel-2 level-2A, which provides 12 multi-spectral channels, and Sentinel-1 SAR, which includes two channels (VV and VH).
\begin{figure}[ht!]
      \centering	   
      \includegraphics[width=0.45\textwidth]{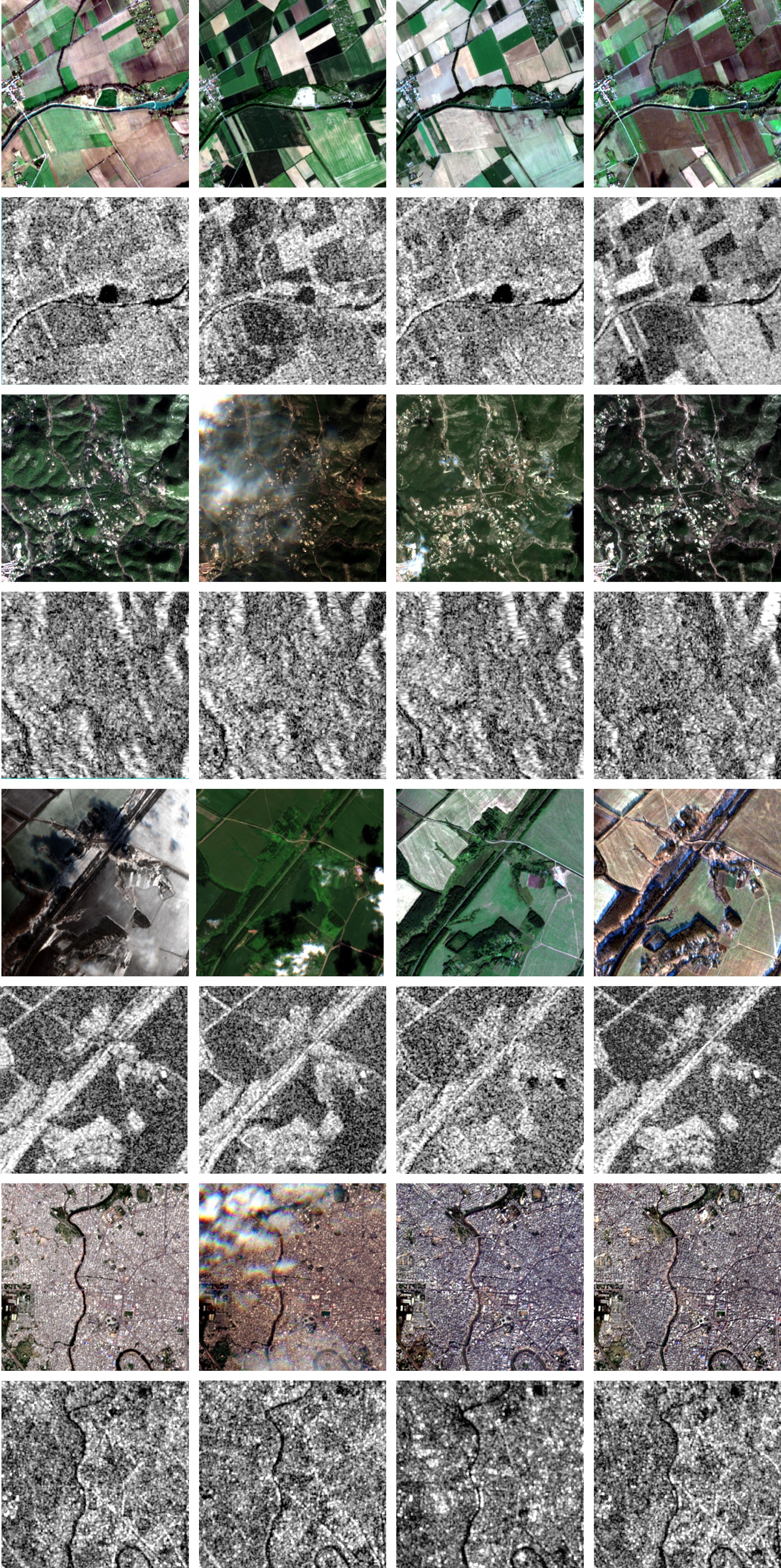}
\caption{\textbf{Sample visualization of the SSL4EO-S12 dataset.} Odd-numbered rows represent Sentinel-2 (multi-spectral), and even-numbered rows represent Sentinel-1 (SAR). Each location is documented with four seasonal snapshots.}
\label{fig:SSL4EO-S12}
\end{figure}
\subsubsection{Pretraining implementation}
We utilize the ViT-Base model as the backbone for our pretraining process. As detailed in Section \ref{Progressive_pretraining_strategy}, our approach incorporates a progressive pretraining strategy, which enhances data utilization and facilitates the learning of representations. Initially, for the single-time point multimodal learning stage, we use only the first season's data, which represents 25\% of the total dataset, and train without the TM block for 20 epochs. In the subsequent multi-time flow multimodal learning stage, we incorporate the TM block and expand pretraining to include the entire dataset throughout 200 epochs. We utilize the computational resources of 8 NVIDIA GeForce RTX 4090 GPUs for the pretraining of our model. To ensure efficient computing, the images are resized to \(128 \times 128\) pixels. Consistent with the normalization methods of the SSL4EO-S12 dataset~\cite{wang2023ssl4eo}, we adopt the same normalization standards. For data augmentation, we apply the partial overlap crop approach mentioned in Section \ref{crop_strategies}, each with random horizontal flipping, as implemented in MAE~\cite{he2022masked}. The batch size is configured at 2048.
The training employs the AdamW optimizer with a base learning rate of \(1\times10^{-4}\), which we modulate using a half-cycle cosine decay schedule over 200 epochs. 
\begin{figure*}
      \centering	   
      \includegraphics[width=1.0\textwidth]{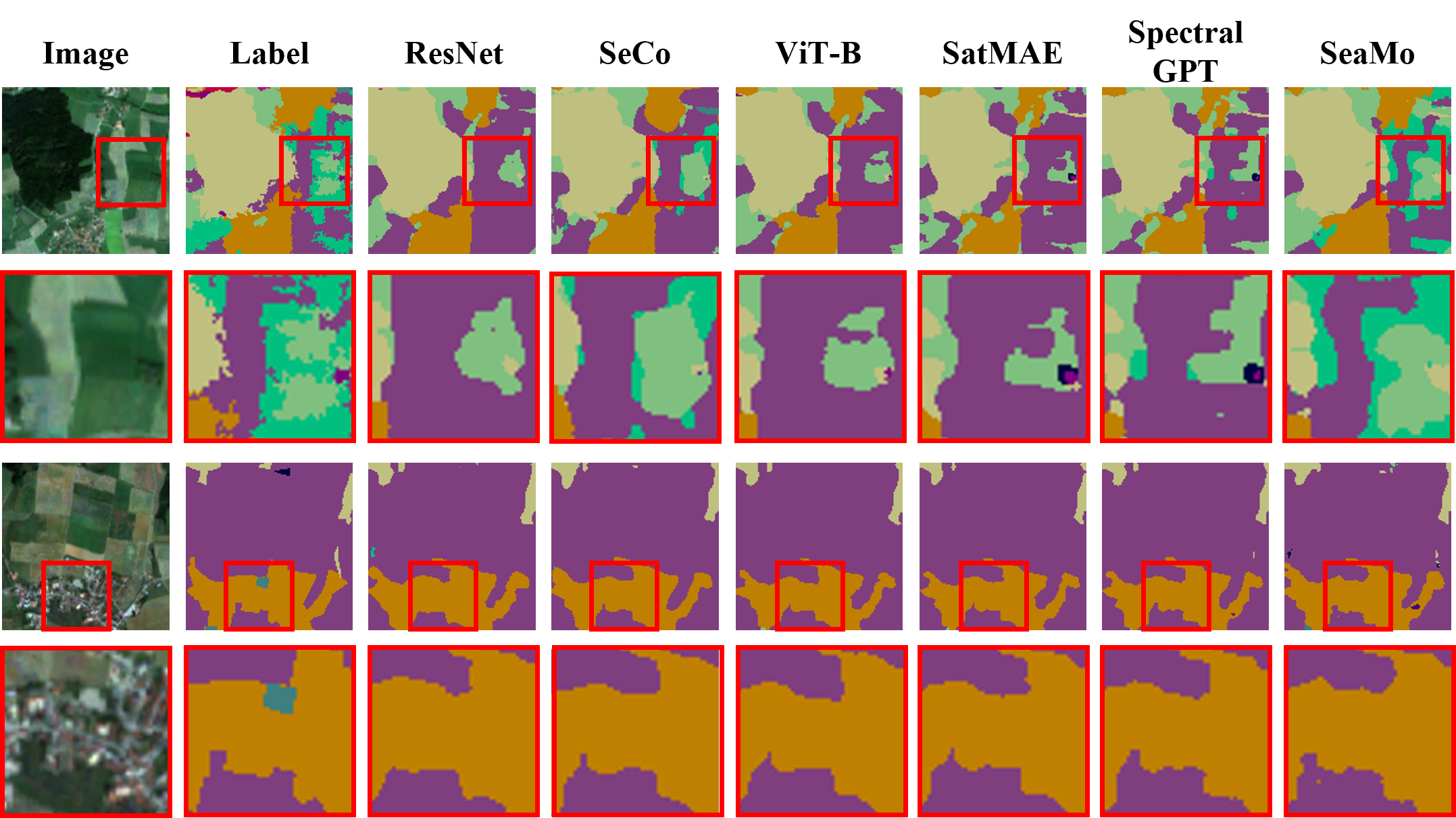}
\caption{\textbf{Visual results obtained by using different pre-trained foundation models for the downstream segmentation task on the SegMunich dataset.}}
\label{fig:SegMunich1}
\end{figure*}
\subsection{Downstream optical experiments}
\label{optical_experiments}
We evaluate our model, SeaMo, across multiple optical datasets, focusing on tasks such as single/multi-label classification, change detection, and semantic segmentation. These datasets vary in terms of scale, image size, sample region, and sensor types. The results demonstrate that our model achieves competitive performance with respect to various benchmarks. To ensure a fair comparison, we select a range of models for evaluation. The compared methods encompass traditional supervised approaches as well as foundation models that utilize different self-supervised methods and are pretrained on different RS datasets.
\subsubsection{Single-label classification} 
We selected the EuroSAT~\cite{helber2019eurosat} and fMoW-Sentinel\cite{cong2022satmae} 
datasets for scene classification, with results displayed in Tables \ref{opt_FT} and \ref{opt_LP}. 

EuroSAT~\cite{helber2019eurosat} is a dataset comprising 27,000 Sentinel-2 satellite images sourced from 34 European countries. These images are divided into 10 scene classification categories, with each category containing between 2,000 and 3,000 labeled images. Each image in the dataset has a resolution of \(64 \times 64\) pixels and includes 13 spectral bands. In line with the methodologies used in SatMAE~\cite{cong2022satmae} and SpectralGPT~\cite{hong2024spectralgpt}, we exclude the cirrus band (B10) and utilize the remaining 12 bands for transfer learning. 
We utilized two NVIDIA GeForce RTX 4090 GPUs for our experiments on the EuroSAT dataset. Images were resized to \(128 \times 128\) pixels. Data augmentation strategies mirrored those used in~\cite{cong2022satmae}, including mixup with a coefficient of 0.8, cutmix of 1.0, and a drop path rate of 0.2. The base learning rate has been set to \(10^{-2}\), with a batch size of 512. The training has been conducted over 150 epochs, including 20 warm-up epochs, and utilizing the AdamW optimizer. The model employed soft cross-entropy loss as the loss function.

The fMoW-Sentinel dataset presents a more challenging scenario with its 63 categories and global sample strategy, resulting in larger performance discrepancies across models. We used only 10\% of the training data for fine-tuning. The input images are resized to \(128 \times 128\) pixels. The base learning rate was set to \(2 \times 10^{-4}\). The training spanned 80 epochs, including 20 warm-up epochs. Data augmentation techniques and loss function implementation have been aligned with those used for the EuroSAT dataset. Remarkably, our model outperforms all others, including those fine-tuned with the full dataset. Notably, though SatMAE, which is pretrained and then fine-tuned on the fMoW-Sentinel, shows better results than other compared models, SeaMo still surpasses SatMAE~\cite{cong2022satmae}, demonstrating superior generalization capabilities of our model.

\subsubsection{Multi-label classification}
We selected the BigEarthNet~\cite{sumbul2019bigearthnet} dataset for multi-label land cover classification. The BigEarthNet (BigEarthNet-MM) dataset~\cite{sumbul2021bigearthnet} consists of 590,326 pairs of Sentinel-1 and Sentinel-2 image patches, designed for multi-modal, multi-label remote sensing image classification. Each image pair is annotated with multi-labels from the CORINE Land Cover (CLC) map of 2018, utilizing the detailed Level-3 class nomenclature. The optical images in the dataset feature 12 multi-spectral channels. In line with practices from SpectralGPT~\cite{hong2024spectralgpt}, we excluded 12\% of the images due to low quality. Consistent with previous studies~\cite{neumann2019domain}, the dataset is divided into 354,196 training samples and 118,065 validation samples. The input images are resized to \(128 \times 128\) pixels. We set the base learning rate to \(10^{-5}\) and maintained a batch size of 512. The training spanned 50 epochs, including 10 warm-up epochs, and employing the AdamW optimizer to optimize our model's performance. All other implementation details remain consistent with those used in the fMoW-Sentinel optical task.
\begin{figure*}[ht!]
      \centering	   
      \includegraphics[width=1.0\textwidth]{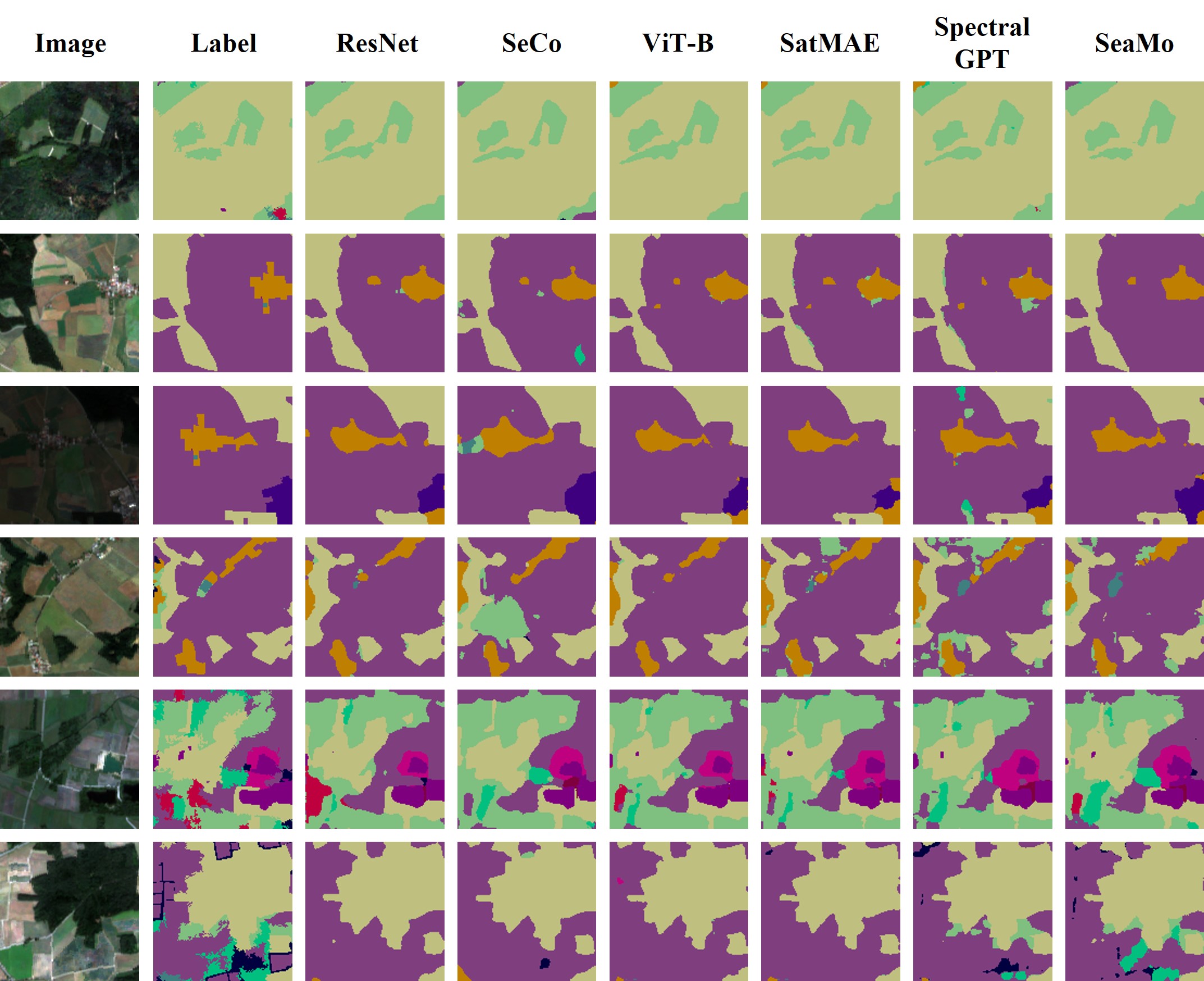}
\caption{\textbf{Visual results obtained by using different pretrained foundation models for the downstream semantic segmentation task on the SegMunich dataset.}}
\label{fig:SegMunich2}
\end{figure*}
In line with the approach used by SpectralGPT~\cite{hong2024spectralgpt}, we fine-tuned our model using only 10\% of the training data. Our model achieved a mean Average Precision (mAP) of 88.54, outperforming all other models. Notably, despite SpectralGPT being pretrained specifically on the BigEarthNet dataset, our model still surpassed its performance, achieving the State-Of-The-Art (SOTA) among all comparable scale models.
\begin{figure*}[ht]
      \centering	   
    \includegraphics[width=1.0\textwidth]{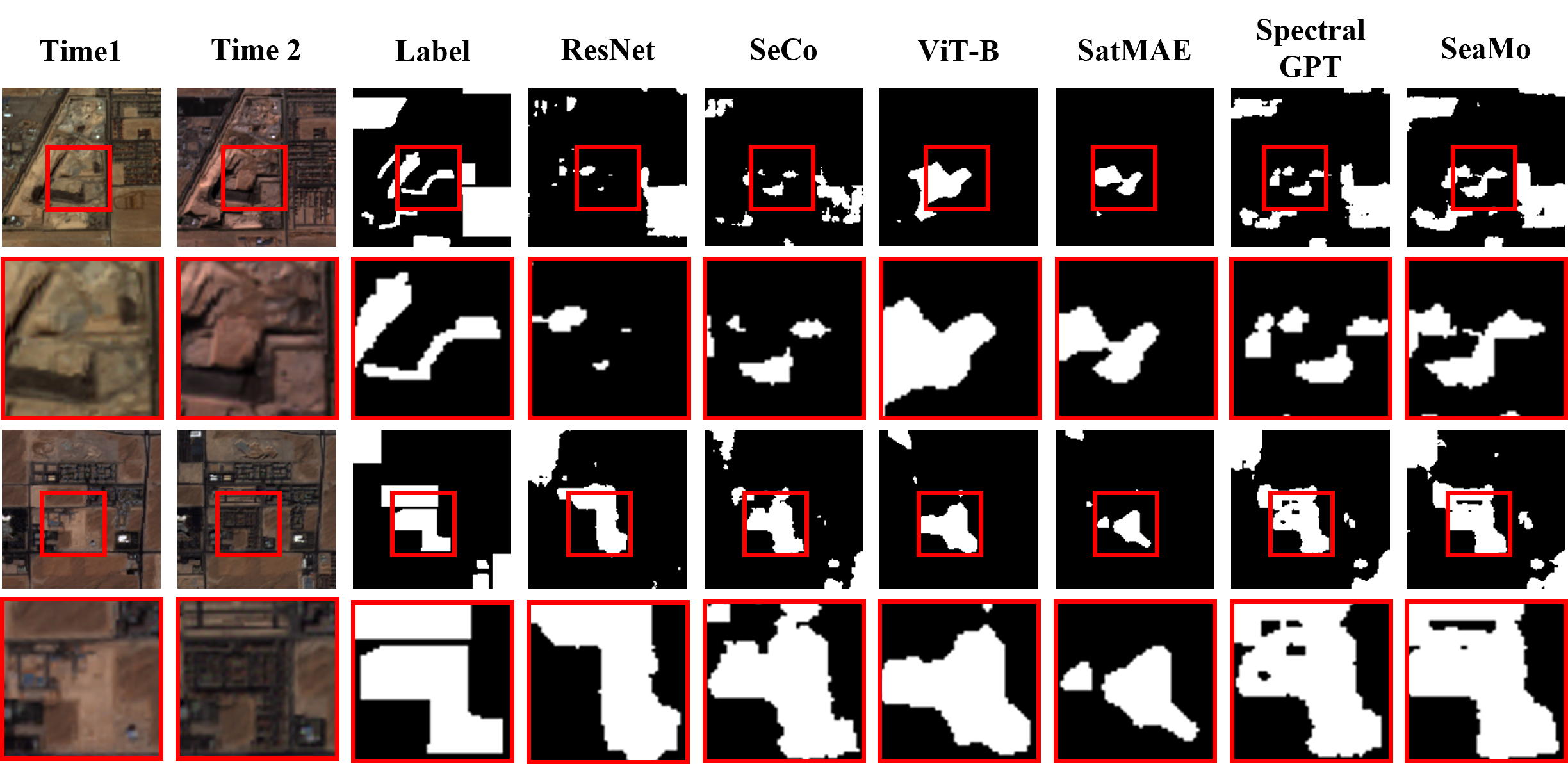}
\caption{\textbf{Visual results obtained by using different pretrained foundation models for the downstream change detection task on the OSCD dataset.}}
\label{fig:OSCD1}
\end{figure*}
\subsubsection{Semantic segmentation}
\begin{table}
\caption{\textbf{FT results on the DFC2020 segmentation dataset.} CL represents the contrastive learning method and MAE represents the masked autoencoder method. Best and second-best results are in red and blue, respectively.}
\centering
\setlength{\tabcolsep}{3pt} 
\label{DFC2020}
\begin{tabular}{cccc}
\toprule
Model &Method & Backbone &mIOU \\
\midrule
\midrule
GASSL~\cite{ayush2021geography}  & CL & ResNet50 & 34.25 \\
SeCo~\cite{manas2021seasonal}  & CL & ResNet50 &36.49 \\
DINO~\cite{caron2021emerging} & CL & ViT-S &32.34 \\
SatMAE~\cite{cong2022satmae}  & MAE &ViT-B & 45.53 \\
SpectralGPT~\cite{hong2024spectralgpt}  & MAE & ViT-B & 44.36 \\
CROMA~\cite{fuller2024croma}  & MAE\&CL &ViT-B & \textbf{\textcolor{blue}{\text{46.67}}} \\
SatMAE~\cite{cong2022satmae}  & MAE & ViT-L & 46.15 \\
\midrule
\textbf{SeaMo} & MAE &ViT-B & \textbf{\textcolor{red}{\text{49.79}}} \\
\bottomrule
\end{tabular}
\end{table}
We selected the DFC2020~\cite{robinson2021global} and SegMunich~\cite{hong2024spectralgpt} datasets to evaluate our model on optical Sentinel-2 segmentation tasks. 

The 2020 IEEE GRSS Data Fusion Contest~\cite{robinson2021global} offers a high-quality global land-cover multi-modal dataset comprising 8 classes. We utilize the dataset provided by Anthony Fuller et al~\cite{fuller2024croma}. This dataset has been preprocessed to include 46,152 training images and 8,874 validation images, with each image resized to \(96 \times 96\) pixels. 
We utilized two NVIDIA GeForce RTX 4090 GPUs for our experiments on the DFC2020 dataset. We maintain the original image size of the dataset at \(96 \times 96\) pixels without resizing. Data augmentation is limited to random vertical and horizontal flips, each with a probability of 0.5. For the optical data task, the base learning rate is set to \(1 \times 10^{-4}\). We fine-tune our model on the optical dataset for 30 epochs, including 5 warm-up epochs. The batch size is set to 64, and we use the soft cross-entropy loss as the loss function.
The results, displayed in Table \ref{DFC2020}, highlight SeaMo's performance, achieving a mean Intersection over Union (mIoU) of 49.79, which is 3.12 points higher than the second-best performing model.
\begin{table}
\caption{\textbf{FT results on the SegMunich segmentation dataset.} CL represents the contrastive learning method and MAE represents the masked autoencoder method. Best and second-best results are in red and blue, respectively.}
\centering
\setlength{\tabcolsep}{3pt} 
\label{SegMunich}
\begin{tabular}{cccc}
\toprule
Model &Method & Backbone &mIOU \\
\midrule
\midrule
GASSL~\cite{ayush2021geography}  & CL & ResNet50 & 45.6 \\
SeCo~\cite{manas2021seasonal}  & CL & ResNet50 &45.9 \\
DINO~\cite{caron2021emerging} & CL & ViT-S &46.5 \\
SatMAE~\cite{cong2022satmae}  & MAE &ViT-B & 48.7 \\
SpectralGPT~\cite{hong2024spectralgpt}  & MAE & ViT-B & 49.8\\
SatMAE~\cite{cong2022satmae}  & MAE & ViT-L & 50.7 \\
SpectralGPT~\cite{hong2024spectralgpt}  & MAE & ViT-L & 51.0\\
CROMA~\cite{fuller2024croma}  & MAE\&CL &ViT-B & \textbf{\textcolor{blue}{\text{51.1}}} \\
\midrule
\textbf{SeaMo} & MAE &ViT-B & \textbf{\textcolor{red}{\text{51.3}}} \\
\bottomrule
\end{tabular}
\end{table}

\begin{figure*}[ht]
      \centering	   
    \includegraphics[width=1.0\textwidth]{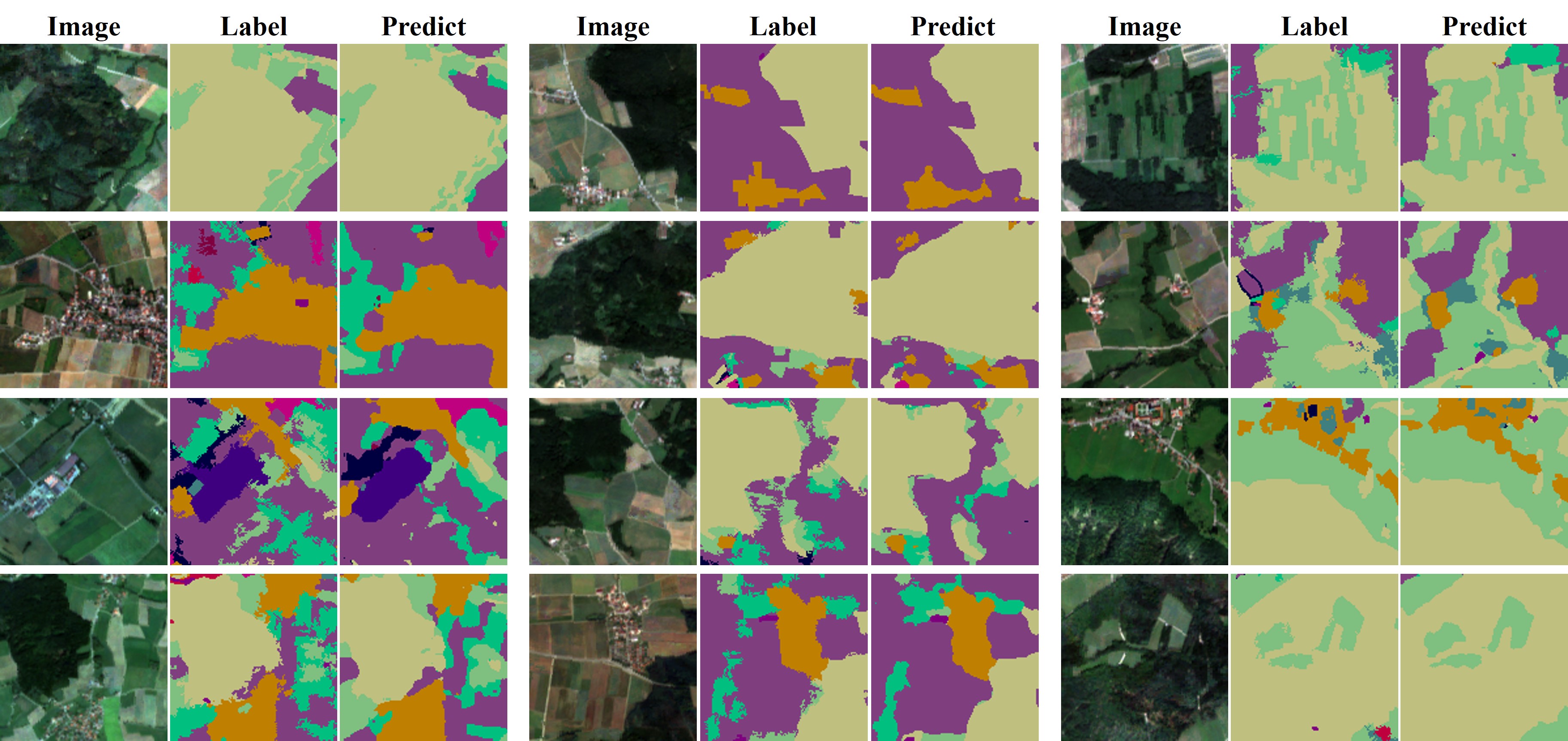}
\caption{\textbf{Additional visualizations of ground truth labels and SeaMo's predicted results on the SegMunich dataset.}}
\label{fig:seg_3}
\end{figure*}

The SegMunich dataset~\cite{hong2024spectralgpt} consists of a 10-band Sentinel-2 image composite, with dimensions of \(3,847 \times 2,958\) pixels and a spatial resolution of 10 meters. It captures the urban landscape of Munich over three years and includes a segmentation mask that accurately delineates 13 Land Use and Land Cover (LULC) classes. We processed the image data into \(128 \times 128\) pixel tokens with a 50\% overlap, following the dataset split introduced in~\cite{hong2024spectralgpt}.
For downstream experiments on the SegMunich dataset, we utilized four NVIDIA GeForce RTX 4090 GPUs. The base learning rate has been set to \(1 \times 10^{-4}\) and the batch size to 128. The model underwent training for 100 epochs to ensure convergence and optimal results. Data augmentation and the loss function were consistent with those used in the DFC2020 tasks. Figures \ref{fig:SegMunich1}, \ref{fig:SegMunich2}, and \ref{fig:seg_3} show representative visualizations illustrating the performance of our model and other comparative methods on the respective datasets. The results are displayed in Table \ref{SegMunich}. SeaMo not only performs exceptionally on this dataset but also achieves SOTA results compared to other models. This demonstrates SeaMo's robust capability in handling complex urban segmentation tasks. 

\subsubsection{Change detection}
\begin{table}
\caption{\textbf{FT results on the OSCD change detection dataset.} CL represents the contrastive learning method and MAE represents the masked autoencoder method. Best and second-best results are in red and blue, respectively.}
\centering
\label{oscd}
\setlength{\tabcolsep}{3pt}
\begin{tabular}{cccc}
\toprule
Model &Method &Backbone &F1 Score \\
\midrule
\midrule
SSL4EO~\cite{wang2023ssl4eo} & CL&ResNet18 & \text{41.85} \\
GASSL~\cite{ayush2021geography}  & CL&ResNet50 & \text{46.26} \\
SeCo~\cite{manas2021seasonal}  & CL&ResNet50 & \text{47.67} \\
CaCO~\cite{mall2023change}  & CL&ResNet50 & \text{52.11} \\
DINO-MC~\cite{wanyan2023dino}  & CL&ViT-S & \text{52.70} \\
ViT-22k~\cite{dosovitskiy2020vit}  & Supervised&ViT-B & \text{52.23} \\
SatMAE~\cite{cong2022satmae}  & MAE&ViT-B & \text{52.76} \\
SpectralGPT~\cite{hong2024spectralgpt}  &MAE& ViT-B & \textbf{\textcolor{blue}{\text{54.29}}} \\
\midrule
\textbf{SeaMo} &MAE & ViT-B & \textbf{\textcolor{red}{\text{54.54}}} \\
\bottomrule
\end{tabular}
\end{table}
Urban change detection is a crucial task in remote sensing applications. 
We chose to evaluate our model using the Onera Satellite Change Detection (OSCD) dataset~\cite{daudt2018urban}. The OSCD dataset comprises 24 pairs of Sentinel-2 images from the period 2015 to 2018.
It includes 14 training and 10 evaluation images, each with 13 spectral bands with resolutions of 10m, 20m, and 60m. Labels within this dataset specify pixel-level urban changes.
\begin{figure*}[ht]
      \centering	   
    \includegraphics[width=1.0\textwidth]{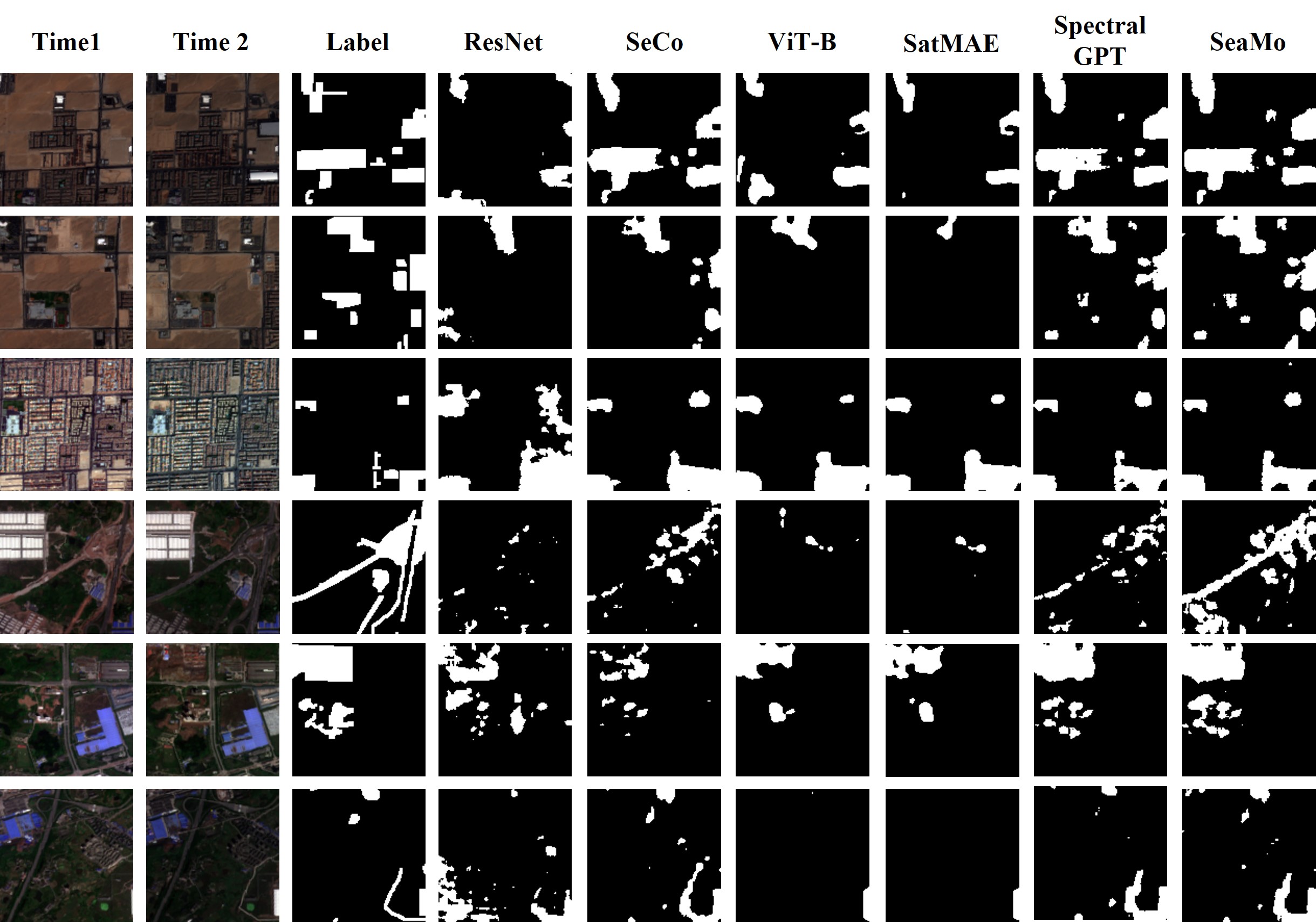}
\caption{\textbf{Visual results obtained by using different pretrained foundation models for the downstream change detection task on the OSCD dataset.}}
\label{fig:OSCD2}
\end{figure*}
Following the approach used in SpectralGPT~\cite{hong2024spectralgpt}, we segmented the image data into \(128 \times 128\) pixel tokens with a 50\% overlap. For the experiments on the OSCD dataset, we utilized four NVIDIA GeForce RTX 4090 GPUs. Data augmentation was limited to random flipping and random rotation to maintain the integrity of the change detection features. The model was trained over 50 epochs, with a batch size of 128 and a learning rate set to \(1 \times 10^{-4}\). We used negative log-likelihood loss (NLLLoss) as the training objective to optimize model performance effectively. According to results presented in Table \ref{oscd}, our model achieved the highest F1-score of 54.54 compared to other models. Figures \ref{fig:OSCD1} and \ref{fig:OSCD2} display representative visualizations illustrating the performance of our model and other comparative methods on the respective datasets.

While SeaMo shows promising results, it is not a SOTA model for this dataset. The reason might be the inadequate employment of the segmentation head and the backbone, and potential issues arising from data imbalance. The complexity of the ViT architecture employed requires significant data to effectively counteract overfitting. Currently, we utilize the UperNet~\cite{xiao2018unified} head, which differentiates between the output features of two urban images for change detection purposes. In future developments, we aim to extensively explore and refine the segmentation head specifically tailored for change detection tasks. This will involve addressing both architectural and data-related challenges to enhance model performance and establish new benchmarks for SOTA results.
\subsection{Downstream radar experiments}
In addition to evaluating SeaMo on optical downstream tasks, we have extended our assessments to include radar-sourced tasks, comparing its performance against several advanced models. These evaluations encompass single/multi-label classification and semantic segmentation tasks. To ensure a fair comparison, we chose models for evaluation that were either pretrained on multimodal or exclusively on radar data. The results confirm that SeaMo also excels in radar-based tasks, demonstrating its versatility and robust performance across various data sources.

\subsubsection{Single/multi-label classification}
\begin{table}[ht!]
\caption{\textbf{FT results on EuroSAT-SAR dataset.} CL represents the contrastive learning method and MAE represents the masked autoencoder method. Best and second-best results are in red and blue, respectively.}
\centering
\label{eurosat_sar}
\setlength{\tabcolsep}{3pt}
\begin{tabular}{cccc}
\toprule
Model &Method &Backbone &Top1 Acc. \\
\midrule
\midrule
SSL4EO-MAE~\cite{wang2023ssl4eo} & MAE&ViT-S & \text{81.05} \\
DINO-MM~\cite{wang2022self}  & CL&ViT-S & \text{85.43} \\
FGMAE~\cite{wang2024feature}  & MAE&ViT-S & \text{85.90} \\
CROMA~\cite{fuller2024croma}  &MAE& ViT-B & \text{88.42} \\
DOFA~\cite{xiong2024neural}  &MAE& ViT-B & \textbf{\textcolor{blue}{\text{88.59}}} \\
\midrule
\textbf{SeaMo} &MAE & ViT-B & \textbf{\textcolor{red}{\text{89.69}}} \\
\bottomrule
\end{tabular}
\end{table}
For this set of tasks, we selected the EuroSAT-SAR~\cite{wang2023feature} and BigEarthNet-SAR~\cite{sumbul2021bigearthnet} datasets. 
Wang et al.~\cite{wang2023feature} collected the SAR modality dataset, which corresponds to the optical multispectral EuroSAT~\cite{helber2019eurosat} dataset. 
We employ the same hyper-parameter settings as those used in the EuroSAT-Optical dataset~\cite{helber2019eurosat}. Table \ref{eurosat_sar} presents the results of our model alongside those of the compared models.
The BigEarthNet-SAR (BEN-SAR) dataset corresponds to the optical dataset and forms part of the BigEarthNet-MultiModal (MM) dataset. As with the optical tasks, we fine-tuned only 10\% of the training data from BigEarthNet-SAR. For tasks on the BigEarthNet-SAR dataset, we set the base learning rate to \(8 \times 10^{-5}\) and kept all other settings consistent with those used in the optical task. Table \ref{ben_sar} presents the results of our model alongside the compared models.

Due to the inherent differences in imaging mechanisms and quality, the performance of all models on SAR data is generally lower than on optical data. 
However, it is evident that our model significantly outperforms others in SAR tasks, achieving substantial improvements. Notably, the performance of SeaMo on BigEarthNet-SAR surpasses even the MAE model fine-tuned with the entire training dataset. These findings underscore that pretraining with multimodal data can enhance model performance on downstream unimodal tasks more effectively than pretraining with unimodal data, even when employing the same self-supervised method.
\begin{table}[ht!]
\caption{\textbf{FT results on the BigEarthNet-SAR dataset.} CL represents the contrastive learning method and MAE represents the masked autoencoder method. We default to using only 10\% of the training data to fine-tune the models. * denotes the results obtained by fine-tuning with all training data, rather than only 10\% of the data. Best and second-best results are in red and blue, respectively.}
\centering
\label{ben_sar}
\setlength{\tabcolsep}{2pt}
\begin{tabular}{cccc}
\toprule
Model &Method &Backbone &mAP \\
\midrule
\midrule
SSL4EO-MAE~\cite{wang2023ssl4eo} & MAE&ViT-S & 74.90/81.30\textsuperscript{*} \\
DINO-MM~\cite{wang2022self}  & CL&ViT-S & 79.50\textsuperscript{*} \\
FGMAE~\cite{wang2024feature}  & MAE&ViT-S & 78.05 \\
SatViT~\cite{fuller2022satvit}& MAE&ViT-B & 75.41\textsuperscript{*} \\
FusMAE~\cite{10642424}& MAE&ViT-B & 75.50\textsuperscript{*} \\
DOFA~\cite{xiong2024neural}& MAE&ViT-B & 81.46 \\
CROMA~\cite{fuller2024croma}  &MAE& ViT-B & \textbf{\textcolor{blue}{\text{81.57}}} \\
\midrule
\textbf{SeaMo} &MAE & ViT-B & \textbf{\textcolor{red}{\text{82.23}}} \\
\bottomrule
\end{tabular}
\end{table}

\begin{table}
\caption{\textbf{FT results on the DFC2020-SAR segmentation dataset.} CL represents the contrastive learning method and MAE represents the masked autoencoder method. Best and second-best results are in red and blue, respectively.}
\centering
\label{dfc_sar}
\setlength{\tabcolsep}{3pt}
\begin{tabular}{cccc}
\toprule
Model &Method &Backbone &mIoU \\
\midrule
\midrule
SSL4EO-MAE~\cite{wang2023ssl4eo} & MAE&ViT-S & 42.35 \\
DINO-MM~\cite{wang2022self}  & CL&ViT-S & 44.86 \\
SatViT~\cite{fuller2022satvit}& MAE&ViT-B & 46.27 \\
CROMA~\cite{fuller2024croma}  &MAE& ViT-B &48.71 \\
DOFA~\cite{xiong2024neural}  &MAE& ViT-B & \textbf{\textcolor{blue}{\text{49.18}}} \\
\midrule
\textbf{SeaMo} &MAE & ViT-B & \textbf{\textcolor{red}{\text{49.54}}} \\
\bottomrule
\end{tabular}
\end{table}
\subsubsection{Semantic segmentation}
We selected the DFC2020-SAR dataset~\cite{robinson2021global} for the SAR-modality segmentation task, which corresponds to the DFC 2020 optical dataset mentioned in Section \ref{optical_experiments}. We maintain the original image size of the dataset at \(96 \times 96\) pixels without resizing. Data augmentation is limited to random vertical and horizontal flips, each with a probability of 0.5. All training settings are the same as those in the optical dataset. The results, presented in Table \ref{dfc_sar}, show that SeaMo's performance on this SAR-modality dataset is comparable to its performance on the corresponding optical dataset, yet it still achieves SOTA with a mIoU of 49.54, which is 0.83 points higher than the second-best model. Additionally, consistent with our earlier observations, models pretrained with multimodal data consistently outperform those pretrained with SAR-only data.
\begin{table}[ht]
\caption{\textbf{FT results on the GLH-Water and S1S2-Water segmentation datasets.} S1S2-Water (S1) denotes fine-tuning performed on Sentinel-1 data, while S1S2-Water (S2) represents fine-tuning on Sentinel-2 data. Best and second-best results are in red and blue, respectively.}
\centering
\label{water_seg}
\setlength{\tabcolsep}{1pt}
\begin{tabular}{ccccccc}
    \toprule
      \multirow{2}{*}{Model}& \multicolumn{2}{c}{GLH-Water} & \multicolumn{2}{c}{S1S2-Water (S1)}  &\multicolumn{2}{c}{S1S2-Water (S2)} \\
         \cmidrule(r){2-3}  \cmidrule(r){4-5}  \cmidrule(r){6-7}
     &IoU&F1&IoU&F1&IoU&F1\\
    \midrule
    \midrule
    UNet~\cite{ronneberger2015u}&82.7&90.2 &87.5&90.2&94.2&96.9\\
    ResNet~\cite{he2016deep}&\textbf{\textcolor{blue}{\text{85.6}}}&\textbf{\textcolor{blue}{\text{92.2}}}&\textbf{\textcolor{blue}{\text{92.5}}}&\textbf{\textcolor{blue}{\text{96.0}}}&97.7&96.5\\
    ViT~\cite{dosovitskiy2020vit}&83.5&90.5&91.6&94.7&\textbf{\textcolor{blue}{\text{98.5}}}&\textbf{\textcolor{blue}{\text{99.3}}}\\
    \midrule
    \textbf{SeaMo}&\textbf{\textcolor{red}{\text{86.2}}}&\textbf{\textcolor{red}{\text{92.8}}}&\textbf{\textcolor{red}{\text{93.7}}}&\textbf{\textcolor{red}{\text{96.7}}}&\textbf{\textcolor{red}{\text{98.7}}}&\textbf{\textcolor{red}{\text{99.5}}}\\
    \bottomrule
\end{tabular}
\end{table}
\begin{figure*}[ht]
      \centering	   
    \includegraphics[width=1.0\textwidth]{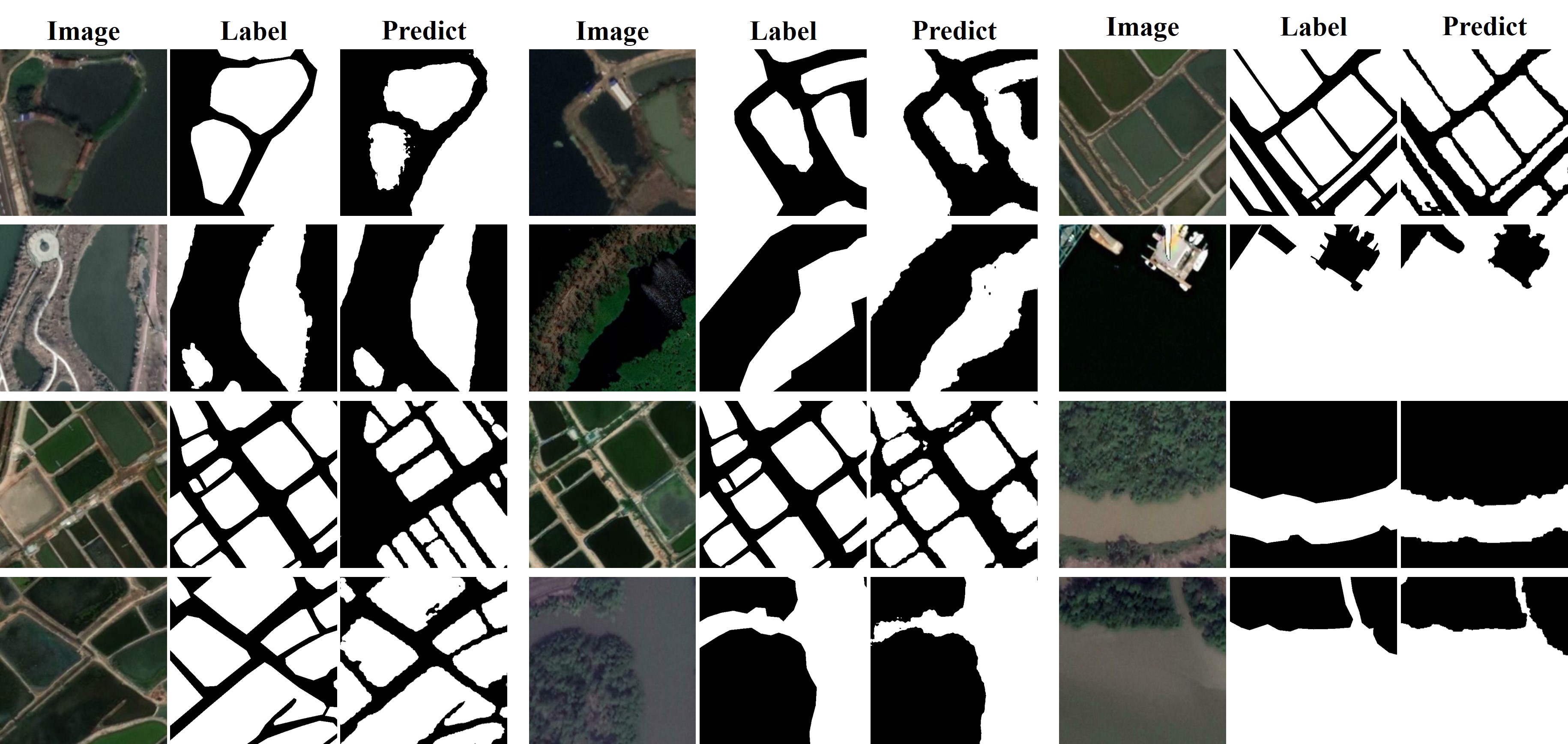}
\caption{\textbf{Visualizations of ground truth labels and SeaMo's predicted results on the GLH-Water dataset.}}
\label{fig:glh_water}
\end{figure*}

\begin{figure*}[ht]
      \centering	   
    \includegraphics[width=1.0\textwidth]{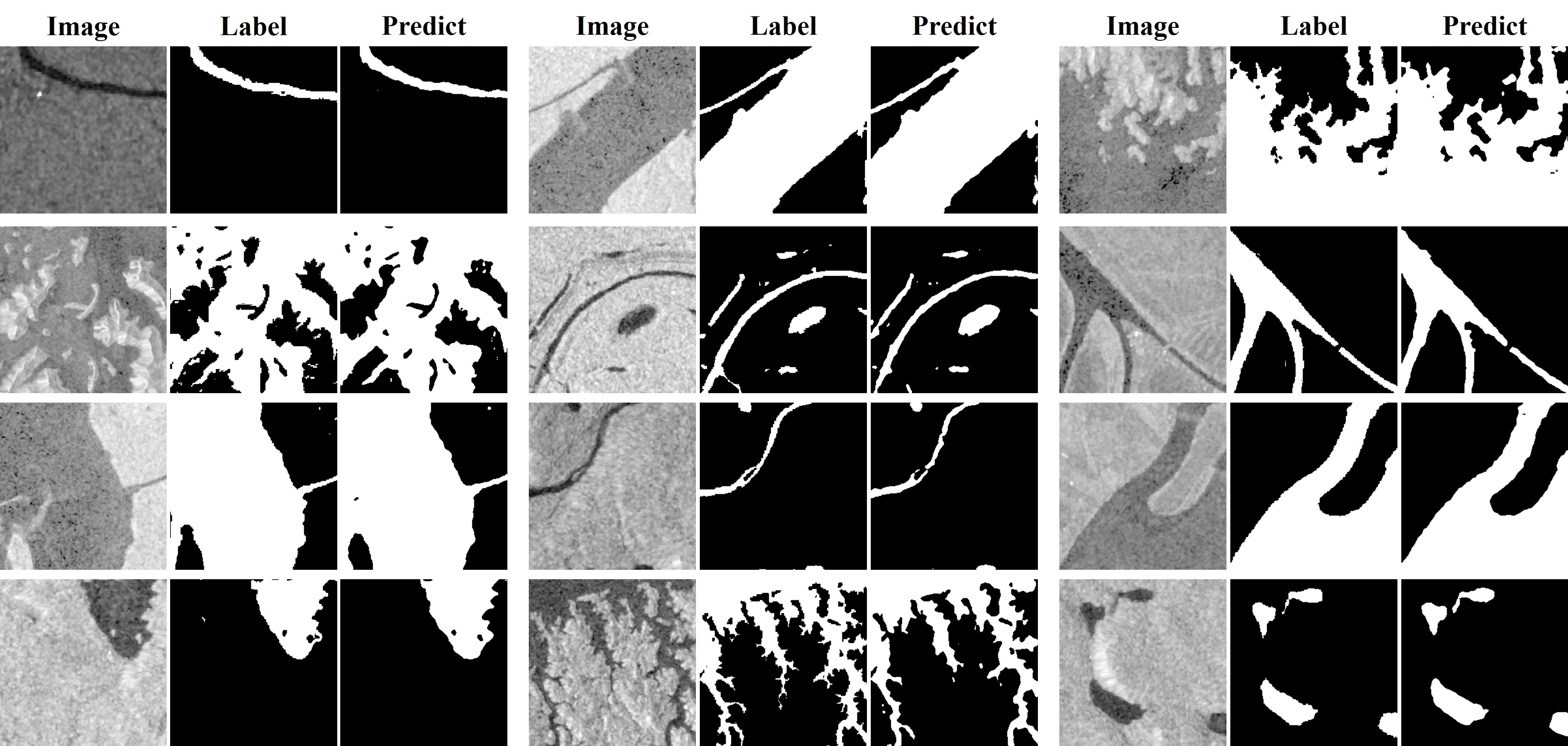}
\caption{\textbf{SAR visual comparisons of ground truth labels and SeaMo's predicted results on the S1S2-Water dataset.} We fine-tuned our model using only the Sentinel-1 data from the dataset.}
\label{fig:s1_water}
\end{figure*}

\begin{figure*}[ht]
      \centering	   
    \includegraphics[width=1.0\textwidth]{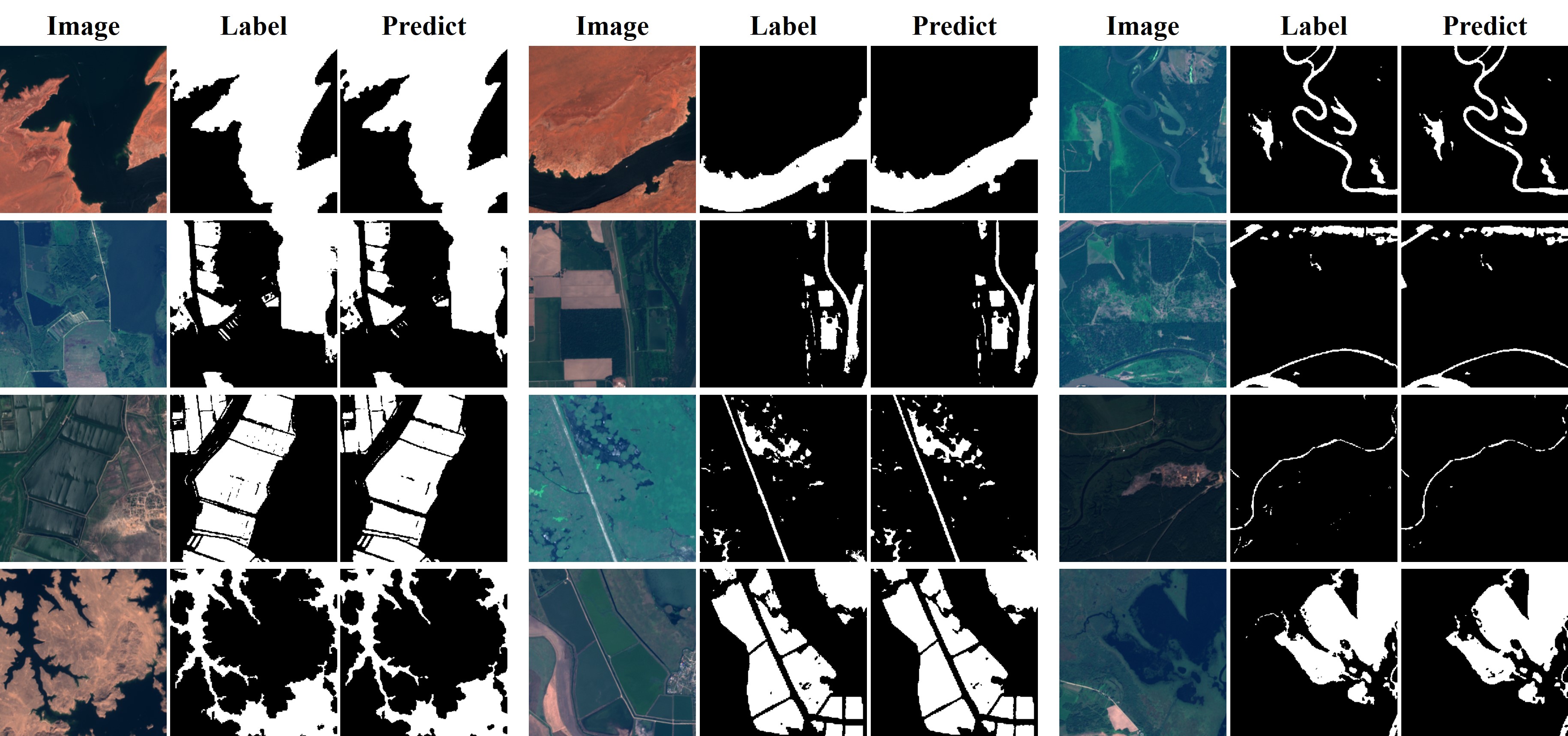}
\caption{\textbf{Optical visual comparisons of ground truth labels and SeaMo's predicted results on the S1S2-Water dataset.} We fine-tuned our model using only the Sentinel-2 data from the dataset.}
\label{fig:s2_water}
\end{figure*}
\begin{figure}[ht!]
      \centering	   
      \includegraphics[width=0.35\textwidth]{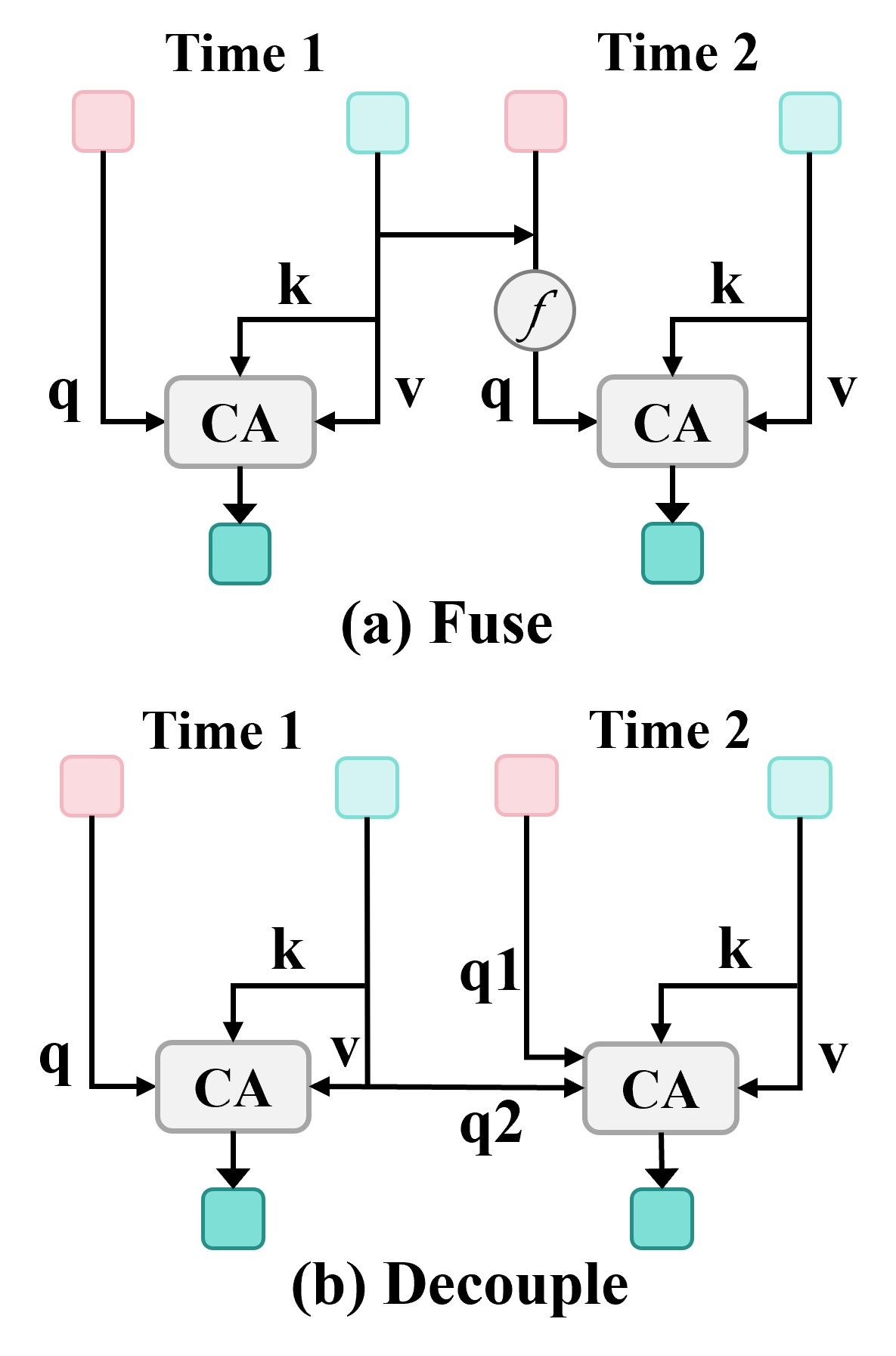}
      \caption{\textbf{Different designs for TM blocks.} (a) \textbf{Fuse}: In this design, a linear layer is used to combine tokens from previous seasons with tokens from the current season. (b) \textbf{Decouple}: In this design, two independent sets of cross-attention layers interact with the modality tokens of the current season. For clarity, the symbols in the figure are defined as follows: $k$ denotes the key vector; $q$, $q1$, and $q2$ denote query vectors; $v$ denotes the value vector; $CA$ represents cross-attention; and $f$ indicates the fully connected layer.}
\label{fig:decouple}
\end{figure}
\begin{figure*}[!t]
      \centering	   
      \includegraphics[width=1.0\textwidth, height=0.3\textwidth]{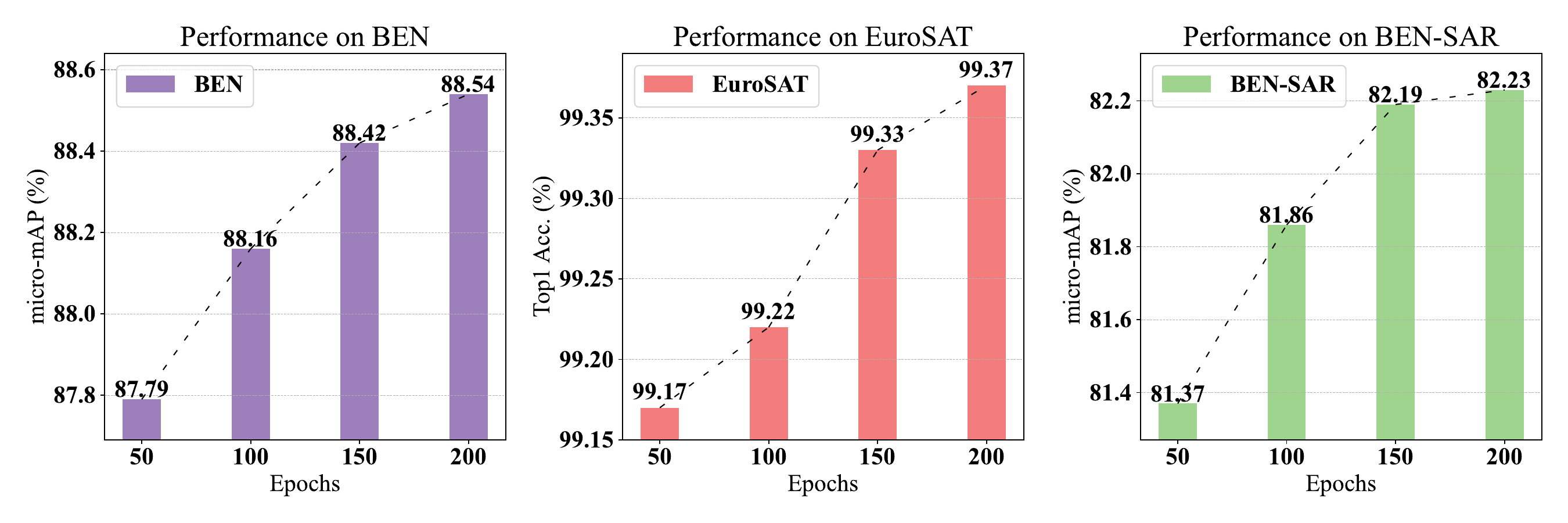}
\caption{\textbf{Pretraining Schedule.} SeaMo was evaluated on the BigEarthNet-optical (BEN), EuroSAT, and BigEarthNet-SAR (BEN-SAR) datasets. Our experiments demonstrate that a longer pretraining duration significantly boosts performance.}
\label{fig:length}
\end{figure*}
\begin{table*}[ht]
\caption{\textbf{Ablations studies of region selection strategies.} Best results are shown in bold and highlighted in green \colorbox[RGB]{217,255,217}{\rule{0pt}{1.5ex}\hspace{1.5ex}}.}
\centering
\label{ablation_crop}
\setlength{\tabcolsep}{5pt}
\begin{tabular}{ccccccc}
    \toprule
      \multirow{3}{*}{Ablation}&\multirow{3}{*}{Crop Rate}& BEN-Optical& EuroSAT-Optical & BEN-SAR  &EuroSAT-SAR &OCSD \\
         \cmidrule(r){3-3} \cmidrule(r){4-4} \cmidrule(r){5-5} \cmidrule(r){6-6} \cmidrule(r){7-7}
     && mAP&Top1 Acc.&Top1 Acc. &Top1 Acc.& F1 Score\\
    \midrule
    \midrule
    Same Location&100\% &\text{87.19} & \text{99.07} & \text{81.57} &87.70 & \text{49.56} \\
    Partial Overlap &[75\%,100\%]&\text{88.40} & \text{99.30} & \text{81.96}& 89.41&\text{53.94}\\
    Partial Overlap & [50\%,100\%]&\cellcolor[RGB]{217,255,217}\textbf{\text{88.54}} & \cellcolor[RGB]{217,255,217}\textbf{\text{99.37}} &\cellcolor[RGB]{217,255,217}\textbf{\text{82.23}} & \cellcolor[RGB]{217,255,217}\textbf{\text{89.69}}&\cellcolor[RGB]{217,255,217}\textbf{\text{54.54}}\\
    Partial Overlap & [25\%,100\%]&\text{88.44} & \text{99.35} &\text{82.10} & 89.55&\text{54.13}\\
    \bottomrule
\end{tabular}
\end{table*}

\begin{table}[ht]
\caption{\textbf{Ablations studies of TM blocks.} Best results are shown in bold and highlighted in green \colorbox[RGB]{217,255,217}{\rule{0pt}{1.5ex}\hspace{1.5ex}}.}
\centering
\label{ablation_TM}
\setlength{\tabcolsep}{3pt}
\begin{tabular}{ccccc}
    \toprule
      & BEN     & EuroSAT & BEN-SAR  & OCSD \\
    \cmidrule(r){2-2}     \cmidrule(r){3-3} \cmidrule(r){4-4} \cmidrule(r){5-5}
     Ablation& mAP&Top1 Acc.&Top1 Acc. & F1 Score\\
    \midrule
    \midrule
    w/o TM Block &\text{88.30} & \text{99.30} & \text{81.74} & \text{53.15} \\
    Decouple &\text{88.46} & \cellcolor[RGB]{217,255,217}\textbf{\text{99.41}} & \cellcolor[RGB]{217,255,217}\textbf{\text{82.30}}&\text{53.94}\\
    Fuse \textbf{(Default)} & \cellcolor[RGB]{217,255,217}\textbf{\text{88.54}} & \text{99.37} &\text{82.23} &\cellcolor[RGB]{217,255,217}\textbf{\text{54.54}}\\
    \bottomrule
\end{tabular}
\end{table}

\subsection{More downstream segmentation tasks}
To further evaluate SeaMo's feature extraction capabilities, we fine-tuned it on more specific remote sensing segmentation datasets focused on single-object recognition tasks. We selected the GLH-Water and S1-S2 Water datasets, dedicated to water body segmentation and encompassing different resolutions and modalities. Table \ref{water_seg} presents the performance of our model alongside the comparison methods on these datasets.

The GLH-Water dataset~\cite{li2024glh} consists of 250 high-resolution satellite images (\(12{,}800 \times 12{,}800\) pixels at 0.3 meters) globally sourced and meticulously annotated for surface water features. These images capture a wide range of water bodies, including rivers, lakes, and ponds against diverse backdrops such as forests, agricultural lands, barren terrains, and urban environments. Following the protocols in the original paper, we cropped these images without overlap, yielding 156,250 segmented images divided into 125,000 for training, 15,625 for validation, and 15,625 for testing. We fine-tuned SeaMo on this dataset over 30 epochs using four NVIDIA GeForce RTX 4090 GPUs, with data augmentation techniques and loss functions consistent with the DFC2020 task, a base learning rate of \(1 \times 10^{-4}\), and a batch size of 24. Figure \ref{fig:glh_water} illustrates SeaMo's segmentation performance on the GLH-Water dataset.

S1-S2 Water~\cite{wieland2023s1s2} is a global reference dataset for semantic segmentation of surface water bodies using publicly available Sentinel-1 and Sentinel-2 images. It comprises 65 triplets of images with quality-checked binary water masks, resulting in over 50,000 non-overlapping \(256 \times 256\) pixel patches for training and over 25,000 patches for validation and testing. Figures \ref{fig:s1_water} and \ref{fig:s2_water} show visualizations of test results after fine-tuning the training sets of Sentinel-1 and Sentinel-2 images. The training settings are the same as those in the GLH-Water implementation. Our model achieves excellent segmentation performance on both the high-resolution GLH-Water dataset and the S1-S2 Water dataset, demonstrating its robustness across different resolutions and modalities.
\subsection{Ablation studies}
\label{ablation_studies}
We conducted ablation studies on SeaMo to delineate the contribution of each component under the default settings: the partial overlap crop within the crop rate from 75\% to 100\%, the fuse type temporal-multimodal block, the Multimodal-Temporal-TM pretraining strategy, a decoder with four blocks, an independent mask ratio of 75\%, reconstruction of normalized patches, and fixed sinusoidal positional embeddings. These studies spanned three tasks across different modalities. Notably, for fine-tuning purposes, we used only 10\% of the training data from the BigEarthNet optical and SAR datasets~\cite{sumbul2021bigearthnet}. 
\subsubsection{Ablation studies of region selection strategies}
\label{Ablation_studies_of_Crop_strategies}
As described in Section \ref{crop_strategies}, spatial cropping of images from different seasons can constrain the model to learn time-invariant spatial information more effectively. In Table \ref{ablation_crop}, we employed various cropping strategies to investigate their impact on model performance. We maintained consistency in all variables except for spatial cropping, using multimodal inputs and incorporating a TM block architecture. In Table \ref{ablation_crop}, \textbf{same location} means that images from all temporal phases are cropped from the same area, implying that, despite different random masking areas, the reconstruction always targets the same image. \textbf{Partial overlap} refers to randomly cropping a certain ratio of a larger image, ensuring that images from different temporal phases partially overlap. As the range of random cropping ratios increases, the overlap becomes more random, and it is even possible to have no overlap at all. The results demonstrate that the pretrained model performs best under our default condition with the TM Block when the overlap between data from different temporal phases is random but still maintains partial overlap. Performance declines when all temporal images are from the same region, which we attribute to the TM Block's cross-attention mechanism allowing the model to easily optimize by reconstructing the same region from images at different times, thus inadequately mining spatial information. When the overlap area is excessively random, model performance slightly decreases. Overall, temporal data should maintain partial overlap rather than complete uniformity, as this increases the difficulty of the reconstruction task and better facilitates the learning of time-invariant spatial properties.

\subsubsection{Ablation studies of Temporal-Multimodal blocks}
\label{Ablations_studies_of_Temporal_Multimodal_blocks}
We also conducted ablation studies on the design of TM blocks, with the results presented in Table \ref{ablation_TM}. As shown, the model without TM blocks performs worse than the model that incorporates them, as further evidenced in Table \ref{ablation}. Within the TM blocks, we compared two designs: \textbf{Decouple} and \textbf{Fuse}. As illustrated in Figure \ref{fig:decouple}, the \textbf{Fuse} design employs a linear layer to merge tokens from previous seasons with tokens from the current season, using a single set of cross-attention layers. In contrast, the \textbf{Decouple} design keeps tokens from previous seasons separate, interacting with the current season's modality tokens via two independent sets of cross-attention layers.

Table \ref{ablation_TM} indicates that both the \textbf{Decouple} and \textbf{Fuse} designs yield improvements on certain downstream tasks. However, given that the \textbf{Decouple} design incurs a higher computational cost due to an increased number of cross-attention layers, more model parameters, and higher data dimensionality during TM block processing, we adopt the \textbf{Fuse} design as the default setting.
\begin{table}
\caption{\textbf{Ablation studies of pretraining strategy.} In this table, colored cells indicate performance changes relative to the baseline: green cells 
\colorbox[RGB]{217,255,217}{\rule{0pt}{1.5ex}\hspace{1.5ex}} and 
\colorbox[RGB]{131,219,212}{\rule{0pt}{1.5ex}\hspace{1.5ex}}
denote improvements, whereas red/pink cells 
\colorbox[RGB]{251,235,234}{\rule{0pt}{1.5ex}\hspace{1.5ex}},
\colorbox[RGB]{255,203,203}{\rule{0pt}{1.5ex}\hspace{1.5ex}},
\colorbox[RGB]{251,200,181}{\rule{0pt}{1.5ex}\hspace{1.5ex}} and 
\colorbox[RGB]{232,137,129}{\rule{0pt}{1.5ex}\hspace{1.5ex}}
indicate degradations.
}
\centering
\label{ablation}
\setlength{\tabcolsep}{2pt}
\begin{tabular}{cccc}
    \toprule
      & BEN     & EuroSAT & BEN-SAR   \\
    \cmidrule(r){2-2}     \cmidrule(r){3-3} \cmidrule(r){4-4}
     Ablation& mAP&Top1 Acc.&Top1 Acc.\\
    \midrule
    \midrule
    Unimodal & \cellcolor[RGB]{232,137,129}\text{-1.23} & \cellcolor[RGB]{251,235,234}\text{-0.09} & \cellcolor[RGB]{232,137,129}\text{-7.33} \\
    Multimodal & \cellcolor[RGB]{251,200,181}\text{-0.73} & \text{-0.06} & \cellcolor[RGB]{251,200,181}\text{-0.99} \\
    Siamese & \cellcolor[RGB]{232,137,129}\text{-1.01} & \cellcolor[RGB]{255,203,203}\text{-0.28} & \cellcolor[RGB]{232,137,129}\text{-2.40} \\
    Siamese-Temporal & \cellcolor[RGB]{232,137,129}\text{-1.31} & \cellcolor[RGB]{255,203,203}\text{-0.22} & \cellcolor[RGB]{131,219,212}\text{+0.45} \\
    Unimodal-Temporal & \cellcolor[RGB]{251,200,181}\text{-0.87} & \cellcolor[RGB]{251,235,234}\text{-0.11} & \cellcolor[RGB]{232,137,129}\text{-1.56} \\
    Multimodal-Temporal & \cellcolor[RGB]{255,203,203}\text{-0.24} & \cellcolor[RGB]{251,235,234}\text{-0.07} & \cellcolor[RGB]{255,203,203}\text{-0.49} \\
    Multimodal-Temporal-TM & \cellcolor[RGB]{217,255,217}\textbf{\text{88.54}} & \cellcolor[RGB]{217,255,217}\textbf{\text{99.37}} & \cellcolor[RGB]{217,255,217}\textbf{\text{82.23}} \\
    \bottomrule
\end{tabular}
\end{table}
\begin{table*}[ht]
    \centering
    \caption{\textbf{Ablation analysis of the proposed SeaMo for various aspects.} Best results are shown in bold and highlighted in green \colorbox[RGB]{217,255,217}{\rule{0pt}{1.5ex}\hspace{1.5ex}}.}
    \setlength\tabcolsep{5pt}  
    \begin{tabularx}{\textwidth}{@{} >{\hsize=.5\hsize}X >{\hsize=.5\hsize}X @{}}  
        \subcaptionbox{Decoder Depth\label{sub_decoderdepth}}[0.48\textwidth]{
            \begin{tabular}{cccc}
                \toprule
                & BEN & EuroSAT & BEN-SAR \\
                \cmidrule(lr){2-2} \cmidrule(lr){3-3} \cmidrule(lr){4-4}
                Blocks & mAP & Top1 Acc. & mAP \\
                \hline\hline
                2 & \text{88.27} & \text{99.26} & \text{81.16} \\
                4 & \text{88.54} & \cellcolor[RGB]{217,255,217}\textbf{\text{99.37}} & \text{82.23} \\
                6 & \cellcolor[RGB]{217,255,217}\textbf{\text{88.58}} & \cellcolor[RGB]{217,255,217}\textbf{\text{99.37}} & \cellcolor[RGB]{217,255,217}\textbf{\text{82.48}} \\
                \bottomrule
            \end{tabular}
        } &
        \subcaptionbox{Mask Ratio\label{sub_maskratio}}[0.48\textwidth]{
            \begin{tabular}{cccc}
                \toprule
                & BEN & EuroSAT & BEN-SAR \\
                \cmidrule(lr){2-2} \cmidrule(lr){3-3} \cmidrule(lr){4-4}
                Ratio & mAP & Top1 Acc. & mAP \\
                \hline\hline
                50\% & \text{88.26} & \text{99.24} & \text{81.45} \\
                75\% & \cellcolor[RGB]{217,255,217}\textbf{\text{88.54}} & \cellcolor[RGB]{217,255,217}\textbf{\text{99.37}} & \cellcolor[RGB]{217,255,217}\textbf{\text{82.23}} \\
                90\% & \text{88.38} & \text{99.22} & \text{81.09} \\
                \bottomrule
            \end{tabular}
        } \\
        \addlinespace[0.5em] 
        \setlength\tabcolsep{2pt}  
        \subcaptionbox{Reconstruction Target\label{sub_ReconstructionTarget}}[0.5\textwidth]{
            \begin{tabular}{cccc}
                \toprule
                & BEN & EuroSAT & BEN-SAR \\
                \cmidrule(lr){2-2} \cmidrule(lr){3-3} \cmidrule(lr){4-4}
                Cases & mAP & Top1 Acc. & mAP \\
                \hline\hline
                w/o norm & \text{87.45} & \text{99.31} & \text{81.25} \\
                w/ norm & \cellcolor[RGB]{217,255,217}\textbf{\text{88.54}} & \cellcolor[RGB]{217,255,217}\textbf{\text{99.37}} & \cellcolor[RGB]{217,255,217}\textbf{\text{82.23}} \\
                \bottomrule
            \end{tabular}
        } &
        \setlength\tabcolsep{2pt}
        \subcaptionbox{Positional Embedding\label{sub_PositionalEmbedding}}[0.5\textwidth]{
            \begin{tabular}{cccc}
                \toprule
                & BEN & EuroSAT & BEN-SAR \\
                \cmidrule(lr){2-2} \cmidrule(lr){3-3} \cmidrule(lr){4-4}
                Type & mAP & Top1 Acc. & mAP \\
                \hline\hline
                learnable & \text{88.34} & \text{99.35} & \text{82.11} \\
                sinusoidal & \cellcolor[RGB]{217,255,217}\textbf{\text{88.54}} & \cellcolor[RGB]{217,255,217}\textbf{\text{99.37}} & \cellcolor[RGB]{217,255,217}\textbf{\text{82.23}} \\
                \bottomrule
            \end{tabular}
        }
    \end{tabularx}
    \label{multi_ablation}
\end{table*}

\subsubsection{Ablations studies of temporal length}
\label{Ablations_studies_of_Temporal_Length}
Table \ref{ablation_TL} demonstrates the impact of temporal length during the pre-training phase on model performance. The temporal length of 1 means that only multimodal data is used for training, without including temporal data or the TM Block. It is obvious that as the time series increases, the model can learn more robust temporal and spatial representations, which is reflected in the improved performance on downstream tasks. This proves the necessity of our dedicated research and design for temporal information.

\begin{table*}[htbp]
\centering
\caption{\textbf{Ablation studies on temporal length.} Best results are shown in bold and highlighted in green \colorbox[RGB]{217,255,217}{\rule{0pt}{1.5ex}\hspace{1.5ex}}.}
\begin{tabular}{ccccc}
    \toprule
      & BEN     & EuroSAT & BEN-SAR  & OCSD \\
    \cmidrule(r){2-2}     \cmidrule(r){3-3} \cmidrule(r){4-4} \cmidrule(r){5-5}
     Temporal Length& mAP&Top1 Acc.&Top1 Acc. & F1 Score\\
    \midrule
    1 &\text{87.81} & \text{99.30} & \text{81.24} & \text{52.10} \\
    2 &\text{87.92} & \text{99.32} & \text{81.59} & \text{53.17} \\
    3 &\text{88.31} &\text{99.33}&\text{82.16}&\text{53.98}\\
    4 \textbf{(Default)} & \cellcolor[RGB]{217,255,217}\textbf{\text{88.54}} & \cellcolor[RGB]{217,255,217}\textbf{\text{99.37}} &\cellcolor[RGB]{217,255,217}\textbf{\text{82.23}} &\cellcolor[RGB]{217,255,217}\textbf{\text{54.54}}\\
    \bottomrule
\end{tabular}
\label{ablation_TL}
\end{table*}

\subsubsection{Pretraining strategy}
Table \ref{ablation} presents the performance of models pretrained with different strategies, with the Siamese and multimodal strategies illustrated in Figure \ref{fig:con_pretraining}. It is challenging to draw a definitive conclusion from these experiments, but we summarize the findings as follows: (1) All multimodal models consistently outperform their unimodal counterparts, suggesting that integrating multimodal data significantly enhances model performance. 
(2) Multimodal-based models generally surpass Siamese-based models across all optical tasks and most SAR tasks. We attribute this to the multimodal-based strategy's effective use of the self-attention mechanism in the encoder, which optimally fuses information across modalities. The only exception is the BEN-SAR task, where the Siamese-temporal model performs best. We hypothesize that the Siamese architecture’s ability to preserve distinct properties of each modality is advantageous, particularly for datasets with lower imaging quality. (3) The inclusion of temporal information consistently benefits all models, corroborating our hypothesis that models equipped to capture time-invariant features can develop more generalized representations. (4) The TM block enhances the feature extraction capability, validating its necessity and effectiveness in our studies.
\subsubsection{Model design ablations}
Table \ref{multi_ablation} and Figure \ref{fig:length} present the results of our ablation studies, which range from model configuration adjustments to modifications in the pretraining schedule. The findings are summarized as follows: 

(1) \textbf{Decoder depth:} A deeper decoder depth notably enhances model performance. However, considering the trade-off between complexity and benefit, we opted for a medium depth of 4 blocks as our default setting. 

(2) \textbf{Mask ratio:} Table \ref{sub_maskratio} shows the impact of varying the mask ratio. Unlike SpectralGPT, SeaMo exhibits decreased performance at a 90\% mask ratio. This decline may be attributed to the already challenging nature of our pretraining pretext task and differences in tokenizer design. 

(3) \textbf{Normalization:} Consistent with practices in MAE and SatMAE, reconstructing normalized patches has proven effective in enabling the encoder to learn more robust representations. 

(4) \textbf{Positional embedding:} Both conventional learnable and fixed positional embeddings show minimal impact on model performance. In future iterations, we plan to explore more advanced embedding techniques to enhance model performance. 

(5) \textbf{Pretraining length:} Figure \ref{fig:length} illustrates how pretraining length affects performance on three downstream tasks. Longer pretraining durations improve model performance, though the rate of improvement gradually diminishes. These ablation studies provide critical insights into the specific aspects of SeaMo that most significantly influence its effectiveness.

\subsubsection{Generalizability test of the pretrained model}
One distinguishing feature of RS images compared to natural images is the diversity in data types. For instance, spectral data captured by the same satellite can vary in image dimensions and the number of spectral bands depending on the preprocessing applied. Given that our model was pretrained on spectral images with dimensions of $128\times128\times12$, we aimed to assess its sensitivity to variations in image size and the number of spectral bands in downstream tasks. As described in Section \ref{headings}, data entering the pretraining model must first pass through a modality-specific tokenizer to compress information. The weights of this tokenizer may not be suitable for different types of data. In our design, we opt for positional encoding interpolation for data of varying sizes, while for data with different channel counts, we introduce a new tokenizer. Although this approach does not utilize the original tokenizer's pretrained weights, the features are still extracted through the pretrained backbone after projecting. Table \ref{Generalizability} displays the performance of our pretrained model after fine-tuning data with different image sizes and spectral band counts. It is evident that while image size has minimal impact on model performance, a reduction in the number of bands significantly decreases performance. Therefore, future work will need to focus on designing a universal tokenizer that accommodates the diverse range of remote sensing data~\cite{xiong2024neural}.
\begin{table}
\caption{\textbf{Generalizability test of the model across different data types in the EuroSAT dataset.} Best results are shown in bold and highlighted in green \colorbox[RGB]{217,255,217}{\rule{0pt}{1.5ex}\hspace{1.5ex}}.}
\centering
\label{Generalizability}
\setlength{\tabcolsep}{3pt}
\begin{tabular}{cccc}
\toprule
Case &Image Size &Spectral Band &Acc. \\
\midrule
\midrule
Case1  &128$\times$128&9&99.20 \\
Case2&128$\times$128&4&99.08 \\
Case3&224$\times$224&12&99.35\\
Case4&224$\times$224&9&99.26\\
\midrule
Default & 128$\times$128&12&\cellcolor[RGB]{217,255,217}\textbf{\text{99.37}}\\
\bottomrule
\end{tabular}
\end{table}
\subsubsection{Computational burden}
\label{compute}
We estimate the computational resources required for both pretraining and fine-tuning our model, SeaMo, with the metrics presented in terms of RTX4090-24GB GPU hours. Detailed estimates are reported in Table \ref{tab_compute}.
\begin{table}[ht]
  \caption{\textbf{Estimated GPU hours used for pretraining and fine-tuning SeaMo.}}
  \setlength{\tabcolsep}{0.3pt}
  \label{tab_compute}
  \centering
  \begin{tabular}{ccc} 
    \toprule
    Task & Dataset & GPU hours \\
    \midrule
    \midrule
    Classification & EuroSAT (Optical\&SAR) & \text{6} \\
    Classification & fMoW Sentinel & \text{20} \\
    Classification & BigEarthNet (Optical\&SAR) & \text{32} \\
    Change Detection & OSCD & \text{2} \\
    Segmentation & SegMunich & \text{32} \\
    Segmentation & DFC2020 (Optical\&SAR) & \text{20} \\
    Pretraining & SSL4EO-S12 & \text{200} \\
    Ablations & SSL4EO-S12 & \text{2050} \\
    \bottomrule
  \end{tabular}
\end{table}

\section{Conclusion}
\label{Conclusion}
In this paper, we introduce SeaMo, a remote sensing foundation model pretrained in a self-supervised manner using a rich array of multi-seasonal and multimodal data. SeaMo is equipped with feature-extracting encoders, TM blocks, and image-reconstruction decoders. Under a progressive MIM framework, SeaMo utilizes space-unaligned heterogeneous satellite images to learn rich multimodal representations and leverages unimodal multi-seasonal satellite images to learn robust time-invariant spatial representations. SeaMo has demonstrated outstanding performance on a variety of downstream tasks, successfully handling several tasks on optical and SAR datasets, showing great potential for broader RS applications. Key insights from our work include: (1) The design of the foundation model architecture is crucial for learning the relationships between multiple RS properties, such as temporal-multimodal and spatial-temporal properties. (2) A progressive self-supervised pretraining approach, which gradually increases data complexity and refines training strategies, enables the model to develop robust representations at various stages. 

Looking ahead, we plan to expand our dataset by collecting more multimodal and multi-source data and to scale up the model to ViT-Large and ViT-Huge configurations. Additionally, we aim to enhance SeaMo's performance by exploring more advanced tokenizer strategies and positional embeddings. We hope this work will offer valuable insights into the development of foundation models for RS imagery.

\section*{Data availability}
\label{Data availability}
The SSL4EO-S12 data that support the findings of this study are openly available in \url{https://mediatum.ub.tum.de/1660427} at \url{https://doi.org/10.1109/MGRS.2023.3281651}.

The EuroSAT data that support the findings of this study are openly available in \url{https://zenodo.org/records/7711810#.ZAm3k-zMKEA} at \url{https://doi.org/10.1109/IGARSS.2018.8519248}.

The BigEarthNet data that support the findings of this study are openly available in \url{https://bigearth.net/} at \url{https://doi.org/10.1109/MGRS.2021.3089174}.

The fMoW-Sentinel data that support the findings of this study are openly available in \url{https://purl.stanford.edu/vg497cb6002} at \url{https://dl.acm.org/doi/10.5555/3600270.3600285}.

The OSCD data that support the findings of this study are openly available in \url{https://ieee-dataport.org/open-access/oscd-onera-satellite-change-detection} at \url{https://doi.org/10.1109/IGARSS.2018.8518015}.

The SegMunich data that support the findings of this study are openly available in \url{https://huggingface.co/datasets/Moonboy12138/SegMunich/blob/main/TUM_128.zip} at \url{https://doi.org/10.1109/TPAMI.2024.3362475}.

The DFC2020 data that support the findings of this study are openly available in \url{https://ieee-dataport.org/competitions/2020-ieee-grss-data-fusion-contest} at \url{https://doi.org/10.1109/JSTARS.2021.3063849}.

The EuroSAT-SAR data that support the findings of this study are openly available in \url{https://huggingface.co/datasets/wangyi111/EuroSAT-SAR} at \url{https://doi.org/10.48550/arXiv.2310.18653}.

The GLH-Water data that support the findings of this study are openly available in \url{https://jack-bo1220.github.io/project/GLH-water.html#gLH-water-Dataset} at \url{https://doi.org/10.1609/aaai.v38i20.30226}.

The S1S2-Water data that support the findings of this study are openly available in \url{https://zenodo.org/records/11278238} at \url{https://dx.doi.org/10.1109/JSTARS.2023.3333969}.

The real data that support the findings of this study are available from the first author upon reasonable request.

\section*{Acknowledgements}
\label{Acknowledgements}
This work was supported by the National Natural Science Foundation of China under Grant 42271350 and also supported by the International Partnership Program of the Chinese Academy of Sciences under Grant No.313GJHZ2023066FN.


\bibliography{mybibfile}

\begin{thebibliography}{10}
\expandafter\ifx\csname url\endcsname\relax
  \def\url#1{\texttt{#1}}\fi
\expandafter\ifx\csname urlprefix\endcsname\relax\def\urlprefix{URL }\fi
\expandafter\ifx\csname href\endcsname\relax
  \def\href#1#2{#2} \def\path#1{#1}\fi

\bibitem{vivone2024deep}
G.~Vivone, L.-J. Deng, S.~Deng, D.~Hong, M.~Jiang, C.~Li, W.~Li, H.~Shen, X.~Wu, J.-L. Xiao, J.~Yao, M.~Zhang, J.~Chanussot, S.~García, A.~Plaza, Deep learning in remote sensing image fusion: Methods, protocols, data, and future perspectives, IEEE Geoscience and Remote Sensing Magazine 13~(1) (2025) 269--310.
\newblock \href {http://dx.doi.org/10.1109/MGRS.2024.3495516} {\path{doi:10.1109/MGRS.2024.3495516}}.

\bibitem{hong2024spectralgpt}
D.~Hong, B.~Zhang, X.~Li, Y.~Li, C.~Li, J.~Yao, N.~Yokoya, H.~Li, P.~Ghamisi, X.~Jia, A.~Plaza, P.~Gamba, J.~A. Benediktsson, J.~Chanussot, {SpectralGPT}: Spectral remote sensing foundation model, IEEE Transactions on Pattern Analysis and Machine Intelligence 46~(8) (2024) 5227--5244.
\newblock \href {http://dx.doi.org/10.1109/TPAMI.2024.3362475} {\path{doi:10.1109/TPAMI.2024.3362475}}.

\bibitem{li2024casformer}
C.~Li, B.~Zhang, D.~Hong, J.~Zhou, G.~Vivone, S.~Li, J.~Chanussot, Casformer: Cascaded transformers for fusion-aware computational hyperspectral imaging, Inf. Fusion 108 (2024) 102408.
\newblock \href {http://dx.doi.org/10.1016/J.INFFUS.2024.102408} {\path{doi:10.1016/J.INFFUS.2024.102408}}.

\bibitem{hong2024multimodal}
D.~Hong, C.~Li, B.~Zhang, N.~Yokoya, J.~A. Benediktsson, J.~Chanussot, Multimodal artificial intelligence foundation models: Unleashing the power of remote sensing big data in earth observation, The Innovation Geoscience 2~(1) (2024) 100055.
\newblock \href {http://dx.doi.org/10.59717/j.xinn-geo.2024.100055} {\path{doi:10.59717/j.xinn-geo.2024.100055}}.

\bibitem{xu2023ai}
Y.~Xu, T.~Bai, W.~Yu, S.~Chang, P.~M. Atkinson, P.~Ghamisi, {AI} security for geoscience and remote sensing: Challenges and future trends, IEEE Geoscience and Remote Sensing Magazine 11~(2) (2023) 60--85.
\newblock \href {http://dx.doi.org/10.1109/MGRS.2023.3272825} {\path{doi:10.1109/MGRS.2023.3272825}}.

\bibitem{bommasani2021opportunities}
R.~Bommasani, D.~A. Hudson, E.~Adeli, R.~Altman, S.~Arora, S.~von Arx, M.~S. Bernstein, J.~Bohg, A.~Bosselut, E.~Brunskill, et~al., On the opportunities and risks of foundation models (2022).
\newblock \href {http://arxiv.org/abs/2108.07258} {\path{arXiv:2108.07258}}.

\bibitem{wanyan2024extending}
X.~Wanyan, S.~Seneviratne, S.~Shen, M.~Kirley, Extending global-local view alignment for self-supervised learning with remote sensing imagery, in: 2024 IEEE/CVF Conference on Computer Vision and Pattern Recognition Workshops (CVPRW), 2024, pp. 2443--2453.
\newblock \href {http://dx.doi.org/10.1109/CVPRW63382.2024.00251} {\path{doi:10.1109/CVPRW63382.2024.00251}}.

\bibitem{wang2022advancing}
D.~Wang, Q.~Zhang, Y.~Xu, J.~Zhang, B.~Du, D.~Tao, L.~Zhang, Advancing plain vision transformer toward remote sensing foundation model, IEEE Transactions on Geoscience and Remote Sensing 61 (2023) 1--15.
\newblock \href {http://dx.doi.org/10.1109/TGRS.2022.3222818} {\path{doi:10.1109/TGRS.2022.3222818}}.

\bibitem{li2024s2mae}
X.~Li, D.~Hong, J.~Chanussot, {S2MAE}: A spatial-spectral pretraining foundation model for spectral remote sensing data, in: 2024 IEEE/CVF Conference on Computer Vision and Pattern Recognition (CVPR), 2024, pp. 27696--27705.
\newblock \href {http://dx.doi.org/10.1109/CVPR52733.2024.02616} {\path{doi:10.1109/CVPR52733.2024.02616}}.

\bibitem{manas2021seasonal}
O.~Mañas, A.~Lacoste, X.~Giró-i Nieto, D.~Vazquez, P.~Rodríguez, {Seasonal Contrast}: Unsupervised pre-training from uncurated remote sensing data, in: 2021 IEEE/CVF International Conference on Computer Vision (ICCV), 2021, pp. 9394--9403.
\newblock \href {http://dx.doi.org/10.1109/ICCV48922.2021.00928} {\path{doi:10.1109/ICCV48922.2021.00928}}.

\bibitem{mall2023change}
U.~Mall, B.~Hariharan, K.~Bala, Change-aware sampling and contrastive learning for satellite images, in: 2023 IEEE/CVF Conference on Computer Vision and Pattern Recognition (CVPR), 2023, pp. 5261--5270.
\newblock \href {http://dx.doi.org/10.1109/CVPR52729.2023.00509} {\path{doi:10.1109/CVPR52729.2023.00509}}.

\bibitem{sun2022ringmo}
X.~Sun, P.~Wang, W.~Lu, Z.~Zhu, X.~Lu, Q.~He, J.~Li, X.~Rong, Z.~Yang, H.~Chang, Q.~He, G.~Yang, R.~Wang, J.~Lu, K.~Fu, Ringmo: A remote sensing foundation model with masked image modeling, IEEE Transactions on Geoscience and Remote Sensing 61 (2023) 1--22.
\newblock \href {http://dx.doi.org/10.1109/TGRS.2022.3194732} {\path{doi:10.1109/TGRS.2022.3194732}}.

\bibitem{cong2022satmae}
Y.~Cong, S.~Khanna, C.~Meng, P.~Liu, E.~Rozi, Y.~He, M.~Burke, D.~Lobell, S.~Ermon, {SatMAE}: Pre-training transformers for temporal and multi-spectral satellite imagery, in: Advances in Neural Information Processing Systems, Vol.~35, Curran Associates, Inc., 2022, pp. 197--211.

\bibitem{he2022masked}
K.~He, X.~Chen, S.~Xie, Y.~Li, P.~Dollár, R.~Girshick, Masked autoencoders are scalable vision learners, in: 2022 IEEE/CVF Conference on Computer Vision and Pattern Recognition (CVPR), 2022, pp. 15979--15988.
\newblock \href {http://dx.doi.org/10.1109/CVPR52688.2022.01553} {\path{doi:10.1109/CVPR52688.2022.01553}}.

\bibitem{reed2023scale}
C.~J. Reed, R.~Gupta, S.~Li, S.~Brockman, C.~Funk, B.~Clipp, K.~Keutzer, S.~Candido, M.~Uyttendaele, T.~Darrell, {Scale-MAE}: A scale-aware masked autoencoder for multiscale geospatial representation learning, in: 2023 IEEE/CVF International Conference on Computer Vision (ICCV), 2023, pp. 4065--4076.
\newblock \href {http://dx.doi.org/10.1109/ICCV51070.2023.00378} {\path{doi:10.1109/ICCV51070.2023.00378}}.

\bibitem{noman2024rethinking}
M.~Noman, M.~Naseer, H.~Cholakkal, R.~M. Anwar, S.~Khan, F.~S. Khan, Rethinking transformers pre-training for multi-spectral satellite imagery, in: 2024 IEEE/CVF Conference on Computer Vision and Pattern Recognition (CVPR), 2024, pp. 27811--27819.
\newblock \href {http://dx.doi.org/10.1109/CVPR52733.2024.02627} {\path{doi:10.1109/CVPR52733.2024.02627}}.

\bibitem{fuller2024croma}
A.~Fuller, K.~Millard, J.~Green, {CROMA}: Remote sensing representations with contrastive radar-optical masked autoencoders, in: Advances in Neural Information Processing Systems, Vol.~36, Curran Associates, Inc., 2023, pp. 5506--5538.

\bibitem{guo2023skysense}
X.~Guo, J.~Lao, B.~Dang, Y.~Zhang, L.~Yu, L.~Ru, L.~Zhong, Z.~Huang, K.~Wu, D.~Hu, H.~He, J.~Wang, J.~Chen, M.~Yang, Y.~Zhang, Y.~Li, {SkySense}: A multi-modal remote sensing foundation model towards universal interpretation for earth observation imagery, in: 2024 IEEE/CVF Conference on Computer Vision and Pattern Recognition (CVPR), 2024, pp. 27662--27673.
\newblock \href {http://dx.doi.org/10.1109/CVPR52733.2024.02613} {\path{doi:10.1109/CVPR52733.2024.02613}}.

\bibitem{bastani2023satlaspretrain}
F.~Bastani, P.~Wolters, R.~Gupta, J.~Ferdinando, A.~Kembhavi, Satlaspretrain: A large-scale dataset for remote sensing image understanding, in: 2023 IEEE/CVF International Conference on Computer Vision (ICCV), 2023, pp. 16726--16736.
\newblock \href {http://dx.doi.org/10.1109/ICCV51070.2023.01538} {\path{doi:10.1109/ICCV51070.2023.01538}}.

\bibitem{srivastava2015unsupervised}
N.~Srivastava, E.~Mansimov, R.~Salakhutdinov, Unsupervised learning of video representations using lstms, in: Proceedings of the 32nd International Conference on International Conference on Machine Learning - Volume 37, ICML'15, JMLR.org, 2015, p. 843–852.

\bibitem{vondrick2016anticipating}
C.~Vondrick, H.~Pirsiavash, A.~Torralba, Anticipating visual representations from unlabeled video, in: 2016 IEEE Conference on Computer Vision and Pattern Recognition (CVPR), 2016, pp. 98--106.
\newblock \href {http://dx.doi.org/10.1109/CVPR.2016.18} {\path{doi:10.1109/CVPR.2016.18}}.

\bibitem{chen2020simple}
T.~Chen, S.~Kornblith, M.~Norouzi, G.~Hinton, A simple framework for contrastive learning of visual representations, in: Proceedings of the 37th International Conference on Machine Learning, Vol. 119 of Proceedings of Machine Learning Research, PMLR, 2020, pp. 1597--1607.

\bibitem{he2020momentum}
K.~He, H.~Fan, Y.~Wu, S.~Xie, R.~Girshick, Momentum contrast for unsupervised visual representation learning, in: 2020 IEEE/CVF Conference on Computer Vision and Pattern Recognition (CVPR), 2020, pp. 9726--9735.
\newblock \href {http://dx.doi.org/10.1109/CVPR42600.2020.00975} {\path{doi:10.1109/CVPR42600.2020.00975}}.

\bibitem{xie2022simmim}
Z.~Xie, Z.~Zhang, Y.~Cao, Y.~Lin, J.~Bao, Z.~Yao, Q.~Dai, H.~Hu, {SimMIM}: a simple framework for masked image modeling, in: 2022 IEEE/CVF Conference on Computer Vision and Pattern Recognition (CVPR), 2022, pp. 9643--9653.
\newblock \href {http://dx.doi.org/10.1109/CVPR52688.2022.00943} {\path{doi:10.1109/CVPR52688.2022.00943}}.

\bibitem{tao2023siamese}
C.~Tao, X.~Zhu, W.~Su, G.~Huang, B.~Li, J.~Zhou, Y.~Qiao, X.~Wang, J.~Dai, Siamese image modeling for self-supervised vision representation learning, in: 2023 IEEE/CVF Conference on Computer Vision and Pattern Recognition (CVPR), 2023, pp. 2132--2141.
\newblock \href {http://dx.doi.org/10.1109/CVPR52729.2023.00212} {\path{doi:10.1109/CVPR52729.2023.00212}}.

\bibitem{bachmann2022multimae}
R.~Bachmann, D.~Mizrahi, A.~Atanov, A.~Zamir, {MultiMAE}: Multi-modal multi-task masked autoencoders, in: Computer Vision -- ECCV 2022, Springer Nature Switzerland, Cham, 2022, pp. 348--367.
\newblock \href {http://dx.doi.org/10.1007/978-3-031-19836-6_20} {\path{doi:10.1007/978-3-031-19836-6_20}}.

\bibitem{feichtenhofer2022masked}
C.~Feichtenhofer, H.~Fan, Y.~Li, K.~He, Masked autoencoders as spatiotemporal learners, in: Proceedings of the 36th International Conference on Neural Information Processing Systems (NeurIPS 2022), NIPS '22, Curran Associates Inc., Red Hook, NY, USA, 2022, pp. 35946--35958.

\bibitem{zhou2023self}
L.~Zhou, H.~Liu, J.~Bae, J.~He, D.~Samaras, P.~Prasanna, Self pre-training with masked autoencoders for medical image classification and segmentation, in: 2023 IEEE 20th International Symposium on Biomedical Imaging (ISBI), 2023, pp. 1--6.
\newblock \href {http://dx.doi.org/10.1109/ISBI53787.2023.10230477} {\path{doi:10.1109/ISBI53787.2023.10230477}}.

\bibitem{nguyen2023climax}
T.~Nguyen, J.~Brandstetter, A.~Kapoor, J.~K. Gupta, A.~Grover, {ClimaX}: a foundation model for weather and climate, in: Proceedings of the 40th International Conference on Machine Learning, ICML'23, JMLR.org, 2023, pp. 25904--25938.

\bibitem{mendieta2023towards}
M.~Mendieta, B.~Han, X.~Shi, Y.~Zhu, C.~Chen, Towards geospatial foundation models via continual pretraining, in: 2023 IEEE/CVF International Conference on Computer Vision (ICCV), 2023, pp. 16760--16770.
\newblock \href {http://dx.doi.org/10.1109/ICCV51070.2023.01541} {\path{doi:10.1109/ICCV51070.2023.01541}}.

\bibitem{tang2023cross}
M.~Tang, A.~Cozma, K.~Georgiou, H.~Qi, Cross-scale mae: a tale of multi-scale exploitation in remote sensing, in: Proceedings of the 37th International Conference on Neural Information Processing Systems, NIPS '23, Curran Associates Inc., Red Hook, NY, USA, 2023, pp. 20054--20066.

\bibitem{ayush2021geography}
K.~Ayush, B.~Uzkent, C.~Meng, K.~Tanmay, M.~Burke, D.~Lobell, S.~Ermon, Geography-aware self-supervised learning, in: 2021 IEEE/CVF International Conference on Computer Vision (ICCV), 2021, pp. 10161--10170.
\newblock \href {http://dx.doi.org/10.1109/ICCV48922.2021.01002} {\path{doi:10.1109/ICCV48922.2021.01002}}.

\bibitem{10726860}
Y.~Wang, C.~M. Albrecht, X.~X. Zhu, Multilabel-guided soft contrastive learning for efficient earth observation pretraining, IEEE Transactions on Geoscience and Remote Sensing 62 (2024) 1--16.
\newblock \href {http://dx.doi.org/10.1109/TGRS.2024.3466896} {\path{doi:10.1109/TGRS.2024.3466896}}.

\bibitem{10504785}
F.~Liu, D.~Chen, Z.~Guan, X.~Zhou, J.~Zhu, Q.~Ye, L.~Fu, J.~Zhou, Remoteclip: A vision language foundation model for remote sensing, IEEE Transactions on Geoscience and Remote Sensing 62 (2024) 1--16.
\newblock \href {http://dx.doi.org/10.1109/TGRS.2024.3390838} {\path{doi:10.1109/TGRS.2024.3390838}}.

\bibitem{10.5555/3666122.3666501}
V.~V. Cepeda, G.~K. Nayak, M.~Shah, {GeoCLIP}: clip-inspired alignment between locations and images for effective worldwide geo-localization, in: Proceedings of the 37th International Conference on Neural Information Processing Systems, NIPS '23, Curran Associates Inc., Red Hook, NY, USA, 2023, pp. 8690--8701.

\bibitem{li2024vision}
X.~Li, C.~Wen, Y.~Hu, Z.~Yuan, X.~X. Zhu, Vision-language models in remote sensing: Current progress and future trends, IEEE Geoscience and Remote Sensing Magazine 12~(2) (2024) 32--66.
\newblock \href {http://dx.doi.org/10.1109/MGRS.2024.3383473} {\path{doi:10.1109/MGRS.2024.3383473}}.

\bibitem{wang2023ssl4eo}
Y.~Wang, N.~A.~A. Braham, Z.~Xiong, C.~Liu, C.~M. Albrecht, X.~X. Zhu, Ssl4eo-s12: A large-scale multimodal, multitemporal dataset for self-supervised learning in earth observation [software and data sets], IEEE Geoscience and Remote Sensing Magazine 11~(3) (2023) 98--106.
\newblock \href {http://dx.doi.org/10.1109/MGRS.2023.3281651} {\path{doi:10.1109/MGRS.2023.3281651}}.

\bibitem{vaswani2017attention}
A.~Vaswani, N.~Shazeer, N.~Parmar, J.~Uszkoreit, L.~Jones, A.~N. Gomez, L.~Kaiser, I.~Polosukhin, Attention is all you need, in: Proceedings of the 31st International Conference on Neural Information Processing Systems, NIPS'17, Curran Associates Inc., Red Hook, NY, USA, 2017, p. 6000–6010.

\bibitem{dosovitskiy2020vit}
A.~Dosovitskiy, L.~Beyer, A.~Kolesnikov, D.~Weissenborn, X.~Zhai, T.~Unterthiner, M.~Dehghani, M.~Minderer, G.~Heigold, S.~Gelly, J.~Uszkoreit, N.~Houlsby, An image is worth 16x16 words: Transformers for image recognition at scale, ICLR.

\bibitem{gupta2023siamese}
A.~Gupta, J.~Wu, J.~Deng, L.~Fei-Fei, Siamese masked autoencoders, in: Proceedings of the 37th International Conference on Neural Information Processing Systems, NIPS '23, Curran Associates Inc., Red Hook, NY, USA, 2023, pp. 40676--40693.

\bibitem{eymael2024efficient}
A.~Eyma{\"e}l, R.~Vandeghen, A.~Cioppa, S.~Giancola, B.~Ghanem, M.~Van~Droogenbroeck, Efficient image pre-training with siamese cropped masked autoencoders, in: Computer Vision -- ECCV 2024, Springer Nature Switzerland, Cham, 2025, pp. 348--366.
\newblock \href {http://dx.doi.org/10.1007/978-3-031-73337-6_20} {\path{doi:10.1007/978-3-031-73337-6_20}}.

\bibitem{xiong2024neural}
Z.~Xiong, Y.~Wang, F.~Zhang, A.~J. Stewart, J.~Hanna, D.~Borth, I.~Papoutsis, B.~L. Saux, G.~Camps-Valls, X.~X. Zhu, Neural plasticity-inspired multimodal foundation model for earth observation (2024).
\newblock \href {http://arxiv.org/abs/2403.15356} {\path{arXiv:2403.15356}}.

\bibitem{caron2021emerging}
M.~Caron, H.~Touvron, I.~Misra, H.~Jegou, J.~Mairal, P.~Bojanowski, A.~Joulin, Emerging properties in self-supervised vision transformers, in: 2021 IEEE/CVF International Conference on Computer Vision (ICCV), 2021, pp. 9630--9640.
\newblock \href {http://dx.doi.org/10.1109/ICCV48922.2021.00951} {\path{doi:10.1109/ICCV48922.2021.00951}}.

\bibitem{assran2023self}
M.~Assran, Q.~Duval, I.~Misra, P.~Bojanowski, P.~Vincent, M.~Rabbat, Y.~LeCun, N.~Ballas, Self-supervised learning from images with a joint-embedding predictive architecture, in: 2023 IEEE/CVF Conference on Computer Vision and Pattern Recognition (CVPR), 2023, pp. 15619--15629.
\newblock \href {http://dx.doi.org/10.1109/CVPR52729.2023.01499} {\path{doi:10.1109/CVPR52729.2023.01499}}.

\bibitem{helber2019eurosat}
P.~Helber, B.~Bischke, A.~Dengel, D.~Borth, Eurosat: A novel dataset and deep learning benchmark for land use and land cover classification, IEEE Journal of Selected Topics in Applied Earth Observations and Remote Sensing 12~(7) (2019) 2217--2226.
\newblock \href {http://dx.doi.org/10.1109/JSTARS.2019.2918242} {\path{doi:10.1109/JSTARS.2019.2918242}}.

\bibitem{sumbul2019bigearthnet}
G.~Sumbul, M.~Charfuelan, B.~Demir, V.~Markl, Bigearthnet: A large-scale benchmark archive for remote sensing image understanding, in: IGARSS 2019 - 2019 IEEE International Geoscience and Remote Sensing Symposium, 2019, pp. 5901--5904.
\newblock \href {http://dx.doi.org/10.1109/IGARSS.2019.8900532} {\path{doi:10.1109/IGARSS.2019.8900532}}.

\bibitem{sumbul2021bigearthnet}
G.~Sumbul, A.~de~Wall, T.~Kreuziger, F.~Marcelino, H.~Costa, P.~Benevides, M.~Caetano, B.~Demir, V.~Markl, Bigearthnet-mm: A large-scale, multimodal, multilabel benchmark archive for remote sensing image classification and retrieval [software and data sets], IEEE Geoscience and Remote Sensing Magazine 9~(3) (2021) 174--180.
\newblock \href {http://dx.doi.org/10.1109/MGRS.2021.3089174} {\path{doi:10.1109/MGRS.2021.3089174}}.

\bibitem{neumann2019domain}
M.~Neumann, A.~S. Pinto, X.~Zhai, N.~Houlsby, In-domain representation learning for remote sensing, in: AI for Earth Sciences Workshop at International Conference on Learning Representations (ICLR), 2020, pp. 1--20.

\bibitem{robinson2021global}
C.~Robinson, K.~Malkin, N.~Jojic, H.~Chen, R.~Qin, C.~Xiao, M.~Schmitt, P.~Ghamisi, R.~Hänsch, N.~Yokoya, Global land-cover mapping with weak supervision: Outcome of the 2020 ieee grss data fusion contest, IEEE Journal of Selected Topics in Applied Earth Observations and Remote Sensing 14 (2021) 3185--3199.
\newblock \href {http://dx.doi.org/10.1109/JSTARS.2021.3063849} {\path{doi:10.1109/JSTARS.2021.3063849}}.

\bibitem{wanyan2023dino}
X.~Wanyan, S.~Seneviratne, S.~Shen, M.~Kirley, {DINO-MC:} self-supervised contrastive learning for remote sensing imagery with multi-sized local crops, CoRR abs/2303.06670.
\newblock \href {http://arxiv.org/abs/2303.06670} {\path{arXiv:2303.06670}}, \href {http://dx.doi.org/10.48550/ARXIV.2303.06670} {\path{doi:10.48550/ARXIV.2303.06670}}.

\bibitem{daudt2018urban}
R.~C. Daudt, B.~Le~Saux, A.~Boulch, Y.~Gousseau, Urban change detection for multispectral earth observation using convolutional neural networks, in: IGARSS 2018 - 2018 IEEE International Geoscience and Remote Sensing Symposium, 2018, pp. 2115--2118.
\newblock \href {http://dx.doi.org/10.1109/IGARSS.2018.8518015} {\path{doi:10.1109/IGARSS.2018.8518015}}.

\bibitem{xiao2018unified}
T.~Xiao, Y.~Liu, B.~Zhou, Y.~Jiang, J.~Sun, Unified perceptual parsing for scene understanding, in: Computer Vision – ECCV 2018: 15th European Conference, Munich, Germany, September 8–14, 2018, Proceedings, Part V, Springer-Verlag, Berlin, Heidelberg, 2018, p. 432–448.
\newblock \href {http://dx.doi.org/10.1007/978-3-030-01228-1_26} {\path{doi:10.1007/978-3-030-01228-1_26}}.

\bibitem{wang2022self}
Y.~Wang, C.~M. Albrecht, X.~X. Zhu, Self-supervised vision transformers for joint sar-optical representation learning, in: IGARSS 2022 - 2022 IEEE International Geoscience and Remote Sensing Symposium, 2022, pp. 139--142.
\newblock \href {http://dx.doi.org/10.1109/IGARSS46834.2022.9883983} {\path{doi:10.1109/IGARSS46834.2022.9883983}}.

\bibitem{wang2024feature}
Y.~Wang, H.~H. Hernández, C.~M. Albrecht, X.~X. Zhu, Feature guided masked autoencoder for self-supervised learning in remote sensing, IEEE Journal of Selected Topics in Applied Earth Observations and Remote Sensing 18 (2025) 321--336.
\newblock \href {http://dx.doi.org/10.1109/JSTARS.2024.3493237} {\path{doi:10.1109/JSTARS.2024.3493237}}.

\bibitem{wang2023feature}
Y.~Wang, C.~M. Albrecht, X.~X. Zhu, Self-supervised vision transformers for joint sar-optical representation learning, in: IGARSS 2022 - 2022 IEEE International Geoscience and Remote Sensing Symposium, 2022, pp. 139--142.
\newblock \href {http://dx.doi.org/10.1109/IGARSS46834.2022.9883983} {\path{doi:10.1109/IGARSS46834.2022.9883983}}.

\bibitem{fuller2022satvit}
A.~Fuller, K.~Millard, J.~R. Green, Satvit: Pretraining transformers for earth observation, IEEE Geoscience and Remote Sensing Letters 19 (2022) 1--5.
\newblock \href {http://dx.doi.org/10.1109/LGRS.2022.3201489} {\path{doi:10.1109/LGRS.2022.3201489}}.

\bibitem{10642424}
H.~Chan-To-Hing, B.~Veeravalli, Fus-mae: A cross-attention-based data fusion approach for masked autoencoders in remote sensing, in: IGARSS 2024 - 2024 IEEE International Geoscience and Remote Sensing Symposium, 2024, pp. 6953--6958.
\newblock \href {http://dx.doi.org/10.1109/IGARSS53475.2024.10642424} {\path{doi:10.1109/IGARSS53475.2024.10642424}}.

\bibitem{ronneberger2015u}
O.~Ronneberger, P.~Fischer, T.~Brox, {U-Net}: Convolutional networks for biomedical image segmentation, in: N.~Navab, J.~Hornegger, W.~M. Wells, A.~F. Frangi (Eds.), Medical Image Computing and Computer-Assisted Intervention -- MICCAI 2015, Springer International Publishing, Cham, 2015, pp. 234--241.
\newblock \href {http://dx.doi.org/10.1007/978-3-319-24574-4\_28} {\path{doi:10.1007/978-3-319-24574-4\_28}}.

\bibitem{he2016deep}
K.~He, X.~Zhang, S.~Ren, J.~Sun, Deep residual learning for image recognition, in: 2016 IEEE Conference on Computer Vision and Pattern Recognition (CVPR), 2016, pp. 770--778.
\newblock \href {http://dx.doi.org/10.1109/CVPR.2016.90} {\path{doi:10.1109/CVPR.2016.90}}.

\bibitem{li2024glh}
Y.~Li, B.~Dang, W.~Li, Y.~Zhang, Glh-water: a large-scale dataset for global surface water detection in large-size very-high-resolution satellite imagery, in: Proceedings of the Thirty-Eighth AAAI Conference on Artificial Intelligence and Thirty-Sixth Conference on Innovative Applications of Artificial Intelligence and Fourteenth Symposium on Educational Advances in Artificial Intelligence, AAAI Press, 2024, pp. 22213--22221.
\newblock \href {http://dx.doi.org/10.1609/aaai.v38i20.30226} {\path{doi:10.1609/aaai.v38i20.30226}}.

\bibitem{wieland2023s1s2}
M.~Wieland, F.~Fichtner, S.~Martinis, S.~Groth, C.~Krullikowski, S.~Plank, M.~Motagh, S1s2-water: A global dataset for semantic segmentation of water bodies from sentinel- 1 and sentinel-2 satellite images, IEEE Journal of Selected Topics in Applied Earth Observations and Remote Sensing 17 (2024) 1084--1099.
\newblock \href {http://dx.doi.org/10.1109/JSTARS.2023.3333969} {\path{doi:10.1109/JSTARS.2023.3333969}}.

\end{thebibliography}

\end{document}